\documentclass[letterpaper]{article} %
\usepackage{aaai25}  %
\usepackage{times}  %
\usepackage{helvet}  %
\usepackage{courier}  %
\usepackage[hyphens]{url}  %
\usepackage{graphicx} %
\usepackage{natbib}  %
\usepackage{caption} %
\usepackage{algorithm}
\usepackage{algorithmic}

\usepackage{graphicx}
\usepackage{amsmath}
\usepackage{amssymb}
\usepackage{booktabs}
\usepackage{makecell}
\usepackage{warpcol}
\usepackage{enumitem}
\usepackage{multirow}
\usepackage{multicol}
\usepackage{mathtools}
\usepackage{xcolor}
\usepackage{bbding}
\usepackage{newfloat}
\usepackage{listings}
\DeclareCaptionStyle{ruled}{labelfont=normalfont,labelsep=colon,strut=off} %
\floatstyle{ruled}
\newfloat{listing}{tb}{lst}{}
\floatname{listing}{Listing}
\def\eg{\textit{e.g.}}
\def\ie{\textit{i.e.}}

\def\etc{\textit{etc.}}

\title{MRStyle: A Unified Framework for Color Style Transfer with\\ Multi-Modality Reference}
\author{
    Jiancheng Huang\textsuperscript{\rm 1}\equalcontrib, Yu Gao\textsuperscript{\rm 1}\equalcontrib, Zequn Jie\textsuperscript{\rm 1}\thanks{Croresponding author.},
    Yujie Zhong\textsuperscript{\rm 1},
    Xintong Han\textsuperscript{\rm 2},
    Lin Ma\textsuperscript{\rm 1}
}
\affiliations{
    \textsuperscript{\rm 1}Meituan Inc.
    \textsuperscript{\rm 2}Huya Inc.
    \\
    jianchenghuang@smail.nju.edu.cn, \{nkugaoyu, zequn.nus, hixintonghan, forest.linma\}@gmail.com, jaszhong@hotmail.com
}

\usepackage{bibentry}
 
\begin{document}

\twocolumn[{%
\renewcommand\twocolumn[1][]{#1}%
\maketitle

\begin{center}
    \centering
    \includegraphics[width=1\linewidth]{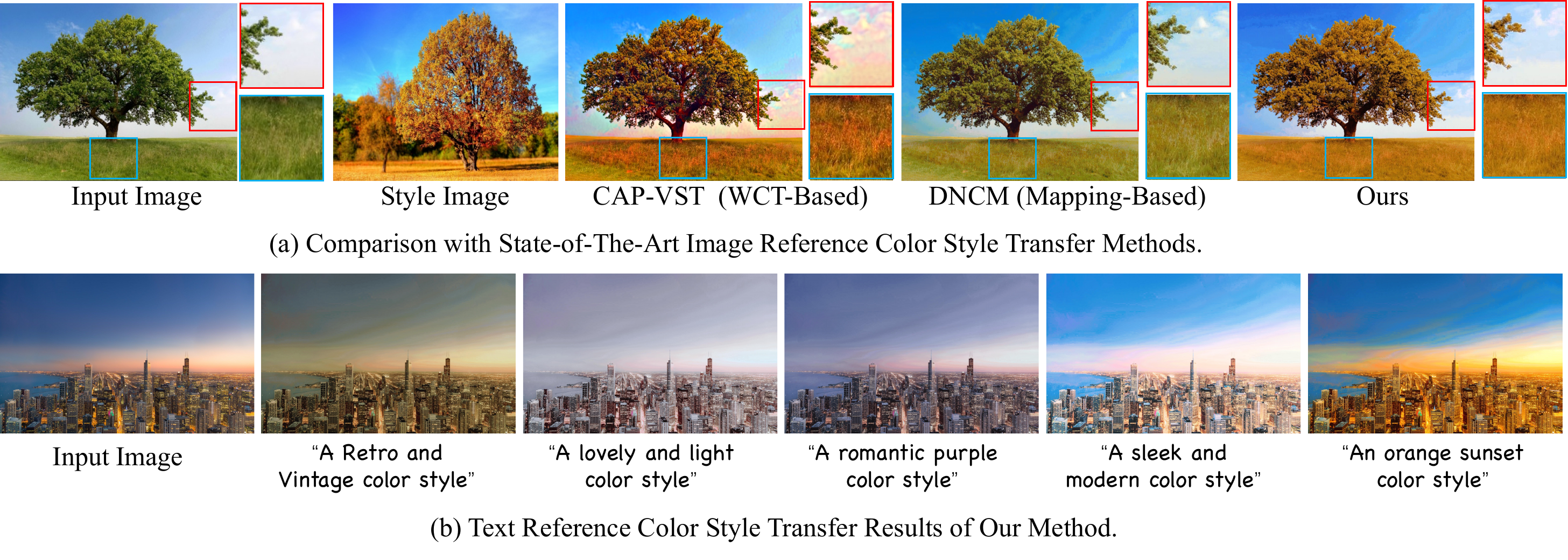}
    \captionof{figure}{
    \textbf{Our multi-modality reference color style transfer results.} (a) State-of-the-art methods DNCM~\cite{ke2023neural} and CAP-VST~\cite{wen2023cap} often produce artifacts (\eg, texture in red box) or unsatisfactory colors (\eg, content in blue box). In contrast, our method avoids artifacts and achieves better color transfer effects.
    (b) Our method can produce amazing stylized results on 8K images given text reference. Zoom in for better visualization.
    }
    \label{fig:quickcomp_new}
\end{center}%
}]
\let\thefootnote\relax\footnotetext{$^*$ Equally contribution \hspace{3pt} $^\dagger$ Corresponding author.}

\begin{abstract}
\vspace{-5pt}
In this paper, we introduce MRStyle, a comprehensive framework that enables color style transfer using multi-modality reference, including image and text. To achieve a unified style feature space for both modalities, we first develop a neural network called IRStyle, which generates stylized 3D lookup tables for image reference. This is accomplished by integrating an interaction dual-mapping network with a combined supervised learning pipeline, resulting in three key benefits: elimination of visual artifacts, efficient handling of high-resolution images with low memory usage, and maintenance of style consistency even in situations with significant color style variations.
For text reference, we align the text feature of stable diffusion priors with the style feature of our IRStyle to perform text-guided color style transfer (TRStyle). Our TRStyle method is highly efficient in both training and inference, producing notable open-set text-guided transfer results.
Extensive experiments in both image and text settings demonstrate that our proposed method outperforms the state-of-the-art in both qualitative and quantitative evaluations.

\end{abstract}

\section{Introduction}
\label{sec:intro}

With the surge in popularity of short video and photo-sharing platforms, many users aspire to customize their photo/video's color style, including brightness, hue, and saturation, before sharing. Existing photo editing software provides expert-defined image filters or lookup tables (LUTs) for color adjustments. However, these pre-set filters/LUTs cannot meet all users' aesthetic needs and limit user flexibility. To mitigate such limitations, researchers have introduced a color-style transfer technology.

Image-guided color style transfer is the most common color style transfer task. It requires the stylized image to align with the reference style image in overall color style while maintaining the content and texture of the original content image. Current image-guided color transfer methods often fall short of fulfilling satisfactory transfer results (Fig.~\ref{fig:quickcomp_new} (a)).
Many recent deep learning-based approaches are built
on encoder-decoder structure with feature whitening and coloring (WCT) operations~\cite{DPST,PhotoWCT,WCT2,an2019ultrafast,DeepPreset,PhotoWCT2}. These methods often produce unrealistic artifacts in the output image due to their reliance on an encoder-decoder structure for stylized image generation. %
Additionally, they have difficulties handling high-resolution images due to the huge network memory usage. An alternative approach, mapping-based methods~\cite{lin2023adacm,ke2023neural}, addresses these issues by applying a predicted color mapping matrix to the original image for color style transfer, instead of using the encoder-decoder pattern. Nevertheless, these methods may fail to deliver satisfactory color transfer effects between images with very different inherent colors, due to the model's structure and training paradigm.

The task of text-guided color style transfer has been recently proposed, as finding a reference style image that meets personal requirements can be challenging and impractical.
Existing text-guided methods~\cite{bau2021paint, patashnik2021styleclip, kwon2022clipstyler, Shi_2022_CVPR} require either expensive paired data gathering or time-consuming online optimization for each content and style. With the development of diffusion models, some image editing methods based on diffusion models can also perform text-guided style transfer~\cite{huang2024diffusion,meng2021sdedit, brooks2022instructpix2pix, huberman2023edit}. However, their stylization quality is not guaranteed as they are not designed for color style transfer.

All these methods mentioned above consider only one modality. Compared to image prompts, text prompts are more user-friendly and flexible but provide less intuitive style information. Consequently, a natural idea springs up, i.e., put forward a unified framework for color style transfer, which can accept either the text or image prompts. The key to this idea is how to unify the style information in the text and image into a common space.
To achieve this, we first train an image-guided color transfer model, and then align the text features from the stable diffusion priors~\cite{rombach2022high} with the color style features of the pre-trained image-guided model. 
\textbf{To the best of our knowledge, our method is the first work that can utilize either image or text prompt as references for color style transfer.}

For image reference, we propose a novel image reference method for color style transfer named IRStyle. Firstly, to avoid artifacts and ensure low memory usage for high-resolution inputs, we follow the mapping-based methods, adopting the simple 3DLUT~\cite{zeng2020lut,cvpr2022Cong} to perform the color transfer. Secondly, to keep style similarity, we introduce an interaction module dual-mapping network.
Additionally, a hybrid training pipeline combining paired supervision and unpaired supervision is designed, which enhances the style similarity metrics.

For text reference, we introduce a lightweight network to align the text features from the pre-trained stable diffusion with the style feature of our IRStyle. 
Since there are no public datasets available for training, we further design a cost-efficient method for data collection with the help of ChatGPT~\cite{ouyang2022training} and the stable diffusion model~\cite{rombach2022high}.
Leveraging the prior of the pre-trained stable diffusion model, our method can conduct open-set text-guided style transfer. In addition, due to the excellent performance of our IRStyle, our method achieves impressive style transfer results (Fig.~\ref{fig:quickcomp_new} (b)). Furthermore, our model structure and data collection strategy ensure high efficiency in both training and testing. The main contributions are summarized as follows:
\begin{itemize}
\setlength{\itemsep}{0pt}
\setlength{\parsep}{0pt}
\setlength{\parskip}{0pt}

\item We propose a generic multi-modality reference color style transfer architecture named MRStyle, which accepts prompts from either images or text as references.
\item For image reference,
we propose a stylized LUTs generation method (IRStyle). Our method can eliminate artifacts, handle high-resolution images with low memory usage, and preserve style uniformity for images with significantly different inherent colors.
\item For text reference, we fully exploit priors from the pre-trained stable diffusion model and our IRStyle to design the text-guided color style transfer model (TRStyle). Our model operates efficiently in both training and inference, as well as generates significant open-set transfer results.
\item Comprehensive evaluations demonstrate that MRStyle outperforms state-of-the-art methods significantly.
\end{itemize}

\section{Related Works}

\noindent\textbf{Image Reference Transfer.}
Image reference color style transfer is the process of color style transition between images. 
This task can be divided into WCT-based methods~\cite{WCT, WCT2, qiao2021sccl, DeepPreset,PhotoWCT, PhotoWCT2, wen2023cap, ijcai2021cGAN} and mapping-based methods~\cite{lin2023adacm, ke2023neural, chen2023nlut}. WCT~\cite{WCT} first uses the feature whitening and coloring operation with an encoder-decoder structure to complete the color transfer. CAP-VST~\cite{wen2023cap} utilizes a similar pipeline with a new effective reversible residual network and an unbiased WCT. Due to the decoding process, they inevitably fall into the problem of visual defects and vast memory requirements. In contrast, mapping-based methods~\cite{lin2023adacm,ke2023neural,chen2023nlut} can well solve these problems, which use low-resolution input to predict the color mapping matrix applied to the original image. AdaCM~\cite{lin2023adacm} proposes a network directly predicting the matrix. DNCM~\cite{ke2023neural} further decouples the process into color normalization and stylization. Nonetheless, the color similarity performance is unsatisfactory. In this work, we propose IRStyle to improve the mapping-based methods, by introducing an interaction dual-mapping network and a combined supervised learning pipeline. In addition to image reference, our method can also utilize text as reference.

\begin{figure*}[t]
\begin{minipage}[b]{1\linewidth}
  \centering
\centerline{\includegraphics[width=1\linewidth]{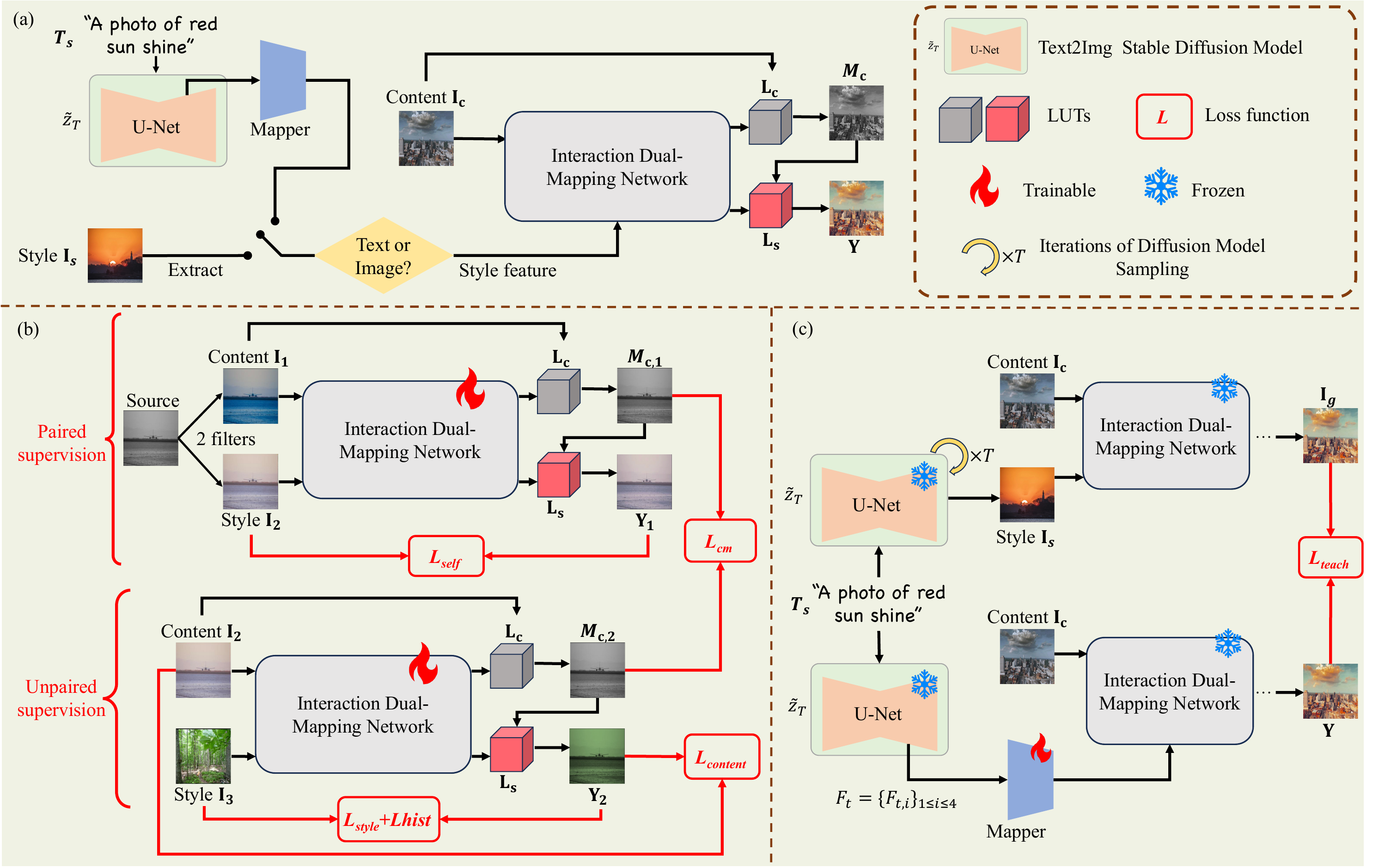}}
\end{minipage}
{\begin{center}
\vspace{-0cm}
\caption{\textbf{The overview of MRStyle.} (a) The inference pipeline of MRStyle. (b) The combined supervised training pipeline of IRStyle. Paired supervised losses are $\mathcal{L}_{self}$ and $\mathcal{L}_{cm}$.
Unpaired supervised losses are $\mathcal{L}_{style}$, $\mathcal{L}_{hist}$ and $\mathcal{L}_{content}$. The Interaction Dual-Mapping Network is detailed in Fig.~\ref{fig:dual-lut} (c). After the training of IRStyle, we integrate the pre-trained IRStyle with stable diffusion priors to finalize our TRStyle. (c) The training pipeline of TRStyle.}
\label{fig:pipeline}
\end{center}
}
\vspace{-0cm}
\end{figure*}

\noindent\textbf{Text Reference Transfer.}
Text reference color style transfer aims at adapting the content image to the color style described by the provided text. 
With the development of the vision-language pre-training models~\cite{radford2021learning, li2022blip, alayrac2022flamingo, singh2022flava}, the information of text and images can be well aligned into a unified space.
SpaceEdit~\cite{Shi_2022_CVPR} performs supervised language-guided image editing, which requires costly paired training data and cannot be applied to open-set scenes.
~\cite{bau2021paint, patashnik2021styleclip, kwon2022clipstyler} have demonstrated the capability of performing open-set text-guided style transfer. They achieve this by leveraging CLIP \cite{radford2021learning} to explore the desired style space during each inference. Nevertheless, this online optimization process incurs significant time consumption during inference, rendering it impractical for real-world applications.
As generative models develop rapidly, image editing, including stylization, has seen considerable improvement. Many image editing methods based on diffusion models are proposed, such as SDEdit~\cite{meng2021sdedit}, EditAnything~\cite{gao2023editanything}, InstructPix2Pix~\cite{brooks2022instructpix2pix}, and MGIE~\cite{fu2023guiding} \etc. However, they primarily concentrate on image editing rather than color style transfer, which can result in content distortion or subpar color outcomes. 
Text-guided colorization\cite{weng2024cad, huang2022unicolor, zabari2023diffusing}, another color-related topic, primarily aims to convert grayscale images into visually pleasing colorful ones, with its text often focusing on object-level color descriptions. Conversely, text-guided color style transfer is primarily concerned with the color style transfer of photorealistic color images, with its text more concentrated on global color style description.
In this study, we exploit how to inject the pre-trained diffusion priors into open-set text-guided color style transfer.

\vspace{-0pt}
\section{Method}
\vspace{-0pt}
The overview of MRStyle is shown in Fig.~\ref{fig:pipeline}.
It can conduct color style transfer using either image or text reference in a unified framework. 
To achieve this, we align the features of reference images and texts into a unified style feature space.
Specifically, we first train IRStyle via a neural LUT network (Sec.\ref{sec:img-ref}). An interaction dual-mapping network and a combined supervised learning pipeline are introduced. Then, we employ the synthetic text-image pairs to train a feature mapper, which projects text features from the pre-trained Stable Diffusion priors into the style space of IRStyle (Sec.~\ref{sec:text-ref}). During inference, when accepting image or text as style input, MRStyle will extract the style features of the corresponding modality, and then interact with the features of the content image for color transfer (Fig.~\ref{fig:pipeline} (a)). 

\vspace{-0pt}
\subsection{Image Reference Color Style Transfer}\label{sec:img-ref}
\vspace{-2pt}
\subsubsection{Interaction Dual-Mapping Network}
\begin{figure*}[t]
\begin{minipage}[b]{1\linewidth}
  \centering
\centerline{\includegraphics[width=\linewidth]{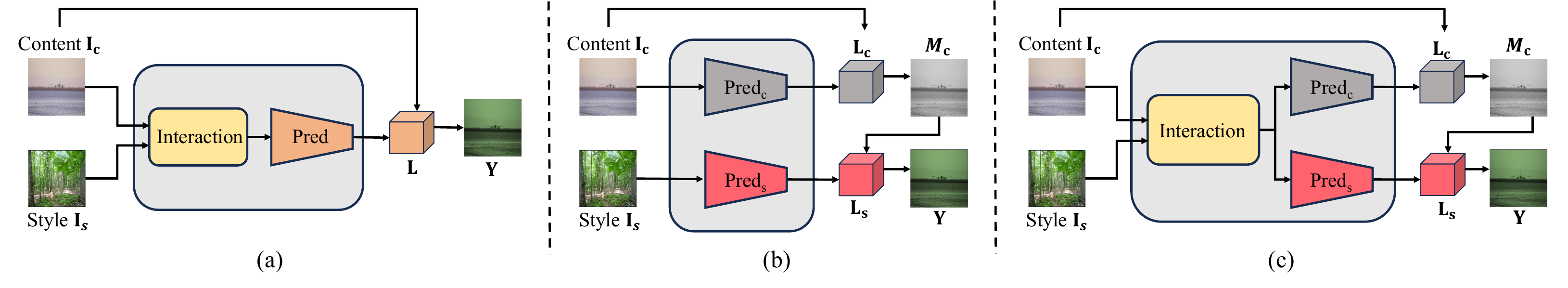}}
\end{minipage}
\vspace{-2pt}
\caption{\textbf{Architectural designs for image reference color style transfer.} The feature extraction of input images has been omitted for simplicity. (a) Interaction direct-mapping network, \eg, AdaCM~\cite{lin2023adacm}. (b) Non-interaction dual-mapping network, \eg, DNCM~\cite{ke2023neural}. (c) Interaction dual-mapping network (ours). }
\label{fig:dual-lut}
\vspace{-1pt}
\end{figure*}

Following mapping-based methods, we design a neural network to generate color mapping matrices and apply them to the original image completing the style color mapping. For simplicity, we consider the combined 3D-LUTs as the mapping matrices. The computational complexity of the 3D-LUT is O(1) for each input pixel, with only 0 floating-point operations, making it extremely fast even at high resolutions. Moreover, when the video scene shows minimal changes, the 3D-LUTs predicted for the initial frame can be applied to subsequent frames, significantly reducing the computational complexity. We posit that our method is not limited to 3D-LUT, other color mapping matrices (e.g. JBL~\cite{JBL}, DNCM~\cite{ke2023neural}) are feasible as well.

Here, we describe how to design our network in detail. Given an original content image $\mathbf{I_c}$ and an reference style image $\mathbf{I_s}$ both with shape $(3, h, w)$, we downsample them to obtain two thumbnail $\mathbf{\tilde{I}_c}$ and $\mathbf{\tilde{I}_s}$.
Then, we feed $\mathbf{\tilde{I}_c}$ and $\mathbf{\tilde{I}_s}$ into a shared encoder $E$ to extract the content features $\mathbf{F_c}$ and style features $\mathbf{F_s}$.
To get the stylized result $\mathbf{Y}$, we compare three possible architectural designs as shown in Fig.~\ref{fig:dual-lut}.

\textbf{a) Interaction direct-mapping network.} As depicted in Fig.~\ref{fig:dual-lut} (a), we simply use an interaction module to merge $\mathbf{F_c}$ and $\mathbf{F_s}$. Then, a predictor $\mathbf{Pred}$ is employed to get a direct transfer LUT $\mathbf{L}$. Finally, we directly map the original image $\mathbf{I_c}$ to the final result $\mathbf{Y}$ through $\mathbf{L}$. This network is similar to AdaCM~\cite{lin2023adacm}. However, directly using a single LUT to do color transfer may be difficult, when the color style between $\mathbf{I_c}$ and $\mathbf{I_s}$ vary largely, as shown in Fig.~\ref{fig:ablation} (e).

\textbf{b) Non-interaction dual-mapping network.} As shown in Fig.~\ref{fig:dual-lut} (b), we employ the content predictor $\mathbf{Pred_c}$ for $\mathbf{F_c}$ to obtain the content LUT $\mathbf{L_c}$, and the style predictor $\mathbf{Pred_s}$ for $\mathbf{F_s}$ to acquire the style LUT $\mathbf{L_s}$. Then we execute a dual mapping (\ie, content extraction and then stylization) to achieve the final outcome. 
First, $\mathbf{L_c}$ is applied on $\mathbf{I_c}$ to get the content map $\mathbf{M_c}$, representing for the style-free content of $\mathbf{I_c}$. 
Second, $\mathbf{L_s}$ is used on $\mathbf{M_c}$ to get the result $\mathbf{Y}$. 
This network is similar to DNCM~\cite{ke2023neural}. 
The prediction of $\mathbf{L_c}$ and $\mathbf{L_s}$ is independent without interaction.
Thus, this requires either the content LUT to normalize all images to a common content space, or the style LUT to transform all contents, which are challenging as shown in Fig.~\ref{fig:ablation} (a). 

\textbf{c) Interaction dual-mapping network.} As previously discussed, we believe that the interaction between content and style features is crucial. 
Furthermore, explicitly decomposing the transfer process into content extraction and stylization can enhance final results.
Thus, we incorporate these two benefits into our final network design as depicted in Fig.~\ref{fig:dual-lut} (c). Specifically, we use the VGG encoder to extract four scales content features $\mathbf{F_c}$ and style features $\mathbf{F_s}$, where $\mathbf{F_s}={\{\mathbf{F_{s,i}}\}_{1\leq i\leq4}}$, $\mathbf{F_c}={\{\mathbf{F_{c,i}}\}_{1\leq i\leq4}}$. These features at multiple scales interact with each other through AdaInt~\cite{huang2017adain}. 
After the interaction, they are downsampled to a uniform scale and concatenated together.
Subsequently, these features are inputted into the content predictor $\mathbf{Pred_c}$ generating the content LUT $\mathbf{L_c}$ of $\mathbf{I_c}$, and also into the style predictor $\mathbf{Pred_s}$ getting the style LUT $\mathbf{L_s}$ of $\mathbf{I_s}$. 
Each predictor is composed of four convolution blocks.
Finally, we apply $\mathbf{L_c}$ to $\mathbf{I_c}$ get the content map $\mathbf{M_c}$, and then utilize $\mathbf{L_s}$ to $\mathbf{M_c}$ get the final result $\mathbf{Y}$. 

\vspace{-4pt}
\subsubsection{The Combined Supervised Learning Pipeline}

A combined learning pipeline is designed to train with paired and unpaired supervision as shown in Fig.~\ref{fig:pipeline} (b). 

Using the paired supervision technique, we randomly apply two filters to a source image, resulting in two images with identical content but different color styles, denoted as $\mathbf{I_1}$ and $\mathbf{I_2}$. Firstly, We take $\mathbf{I_1}$ as the content image and $\mathbf{I_2}$ as the style image. The desired stylized result $\mathbf{Y_1}$ of $\mathbf{I_1}$ should be $\mathbf{I_2}$ itself. Hence MSE loss between $\mathbf{Y_1}$ and $\mathbf{I_2}$ is adopted, referred to as $\mathcal{L}_{self}$.
Secondly, we use $\mathbf{I_1}$ and $\mathbf{I_2}$ as content images for model training respectively. Images with the same content should extract identical content maps, \ie, the content map $\mathbf{M_{c,1}}$ of $\mathbf{I_1}$ and the content map $\mathbf{M_{c,2}}$ of $\mathbf{I_2}$ should be identical.
MSE loss between  $\mathbf{M_{c,1}}$ and  $\mathbf{M_{c,2}}$ is used, denoted as $\mathcal{L}_{cm}$.
The paired supervised losses are:
\begin{equation}\label{eq:Lself}
    \mathcal{L}_{pair} = \mathcal{L}_{self}(\mathbf{Y_1},\mathbf{I_2}) + \mathcal{L}_{cm}(\mathbf{M_{c,1}},\mathbf{M_{c,2}})
\end{equation}

\vspace{-4pt}
The benefit of paired supervision technology lies in its capacity to obtain ground truth in a self-supervised manner. However, it is inconsistent with the real inference situation, where there is typically a difference in content between the style and content image. To address this, we introduce unpaired supervision, where the content image $\mathbf{I_2}$ and style image $\mathbf{I_3}$ are derived from different images. Since there is no ground truth for unpaired supervision, 
we utilize loss functions commonly employed in other color transfer methods~\cite{WCT,WCT2,PhotoWCT,PhotoWCT2}, including style similarity loss $\mathcal{L}_{style}$ and structural content loss $\mathcal{L}_{content}$. $\mathcal{L}_{style}$ is implemented through MSE between the mean and standard deviation of shallow feature maps extracted from pre-trained VGG-Net, while $\mathcal{L}_{content}$ is implemented by MSE between the deep feature maps. However, $\mathcal{L}_{style}$ generally reflects the similarity of the high-level semantic feature space rather than color information~\cite{ke2023neural}. Therefore, we additionally employ a more interpretable loss function $\mathcal{L}_{hist}$ following \cite{huang2023composer}, which measures the distance between the soft color histograms. We denote the resulting image as $\mathbf{Y_2}$. The unpaired supervised losses are as follows:
\vspace{-3pt}
\begin{equation}\label{eq:Lother}
        \mathcal{L}_{unpair} =  
        \mathcal{L}_{content}(\mathbf{Y_2},\mathbf{I}_{2}) +
        \mathcal{L}_{style}(\mathbf{Y_2},\mathbf{I}_{3}) + \mathcal{L}_{hist}(\mathbf{Y_2},\mathbf{I}_{3})
\end{equation}
In training, both paired supervised losses and unpaired supervised losses are utilized for each sample.

\vspace{-1pt}
\subsection{Text Reference Color Style Transfer}\label{sec:text-ref}
\vspace{-1pt}
\subsubsection{A Vanilla Text-Guided Way}\label{sec:naive}

With the fast development of text-to-image diffusion models, the stable diffusion model can produce high-quality images according to the text prompt. Therefore, a vanilla way to achieve text reference color style transfer would be directly using the pre-trained stable diffusion model to generate reference style images by the provided style text. Then, given the generated style image, we can use our pre-trained IRStyle to get the result.

Although this solution can accomplish the task, it has two main drawbacks. Firstly, it is time-consuming as the generation of a style image requires multiple steps within the diffusion process. Secondly, the entire process of text-guided color transfer is not end-to-end. Thus, a faster and more elegant alternative needs to be presented.

\begin{figure}[t]
\begin{minipage}[b]{1\linewidth}
  \centering
\centerline{\includegraphics[width=1\linewidth]{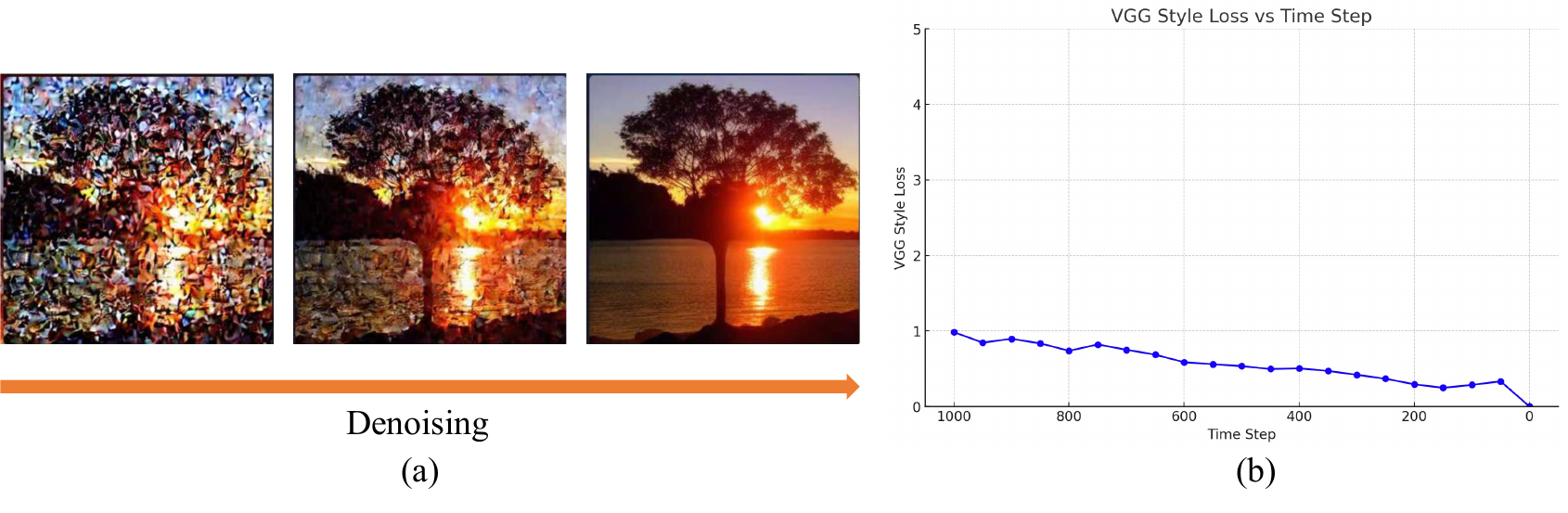}}
\end{minipage}

{\begin{center}
\vspace{-0.1cm}
\caption{\textbf{The color and illumination of the generated image in the diffusion model.} 
(a) The denoising process during image generation. (b) The color similarity analysis under different time steps of diffusion model sampling.}
\label{fig:early}
\end{center}
}
\vspace{-1pt}
\end{figure}

\vspace{-1pt}
\subsubsection{Efficient Priors Feature Mapper}
\label{sec:mapper}
The Stable Diffusion model~\cite{rombach2022high} accomplishes text-to-image generation via a U-Net structure, characterized by a step-by-step denoising process. To ensure that the final generated image is semantically consistent with the given text, the model computes the cross-attention between the text embeddings and U-Net features at every step. This suggests that the internal representations of the U-Net features could be well-associated with language-describable semantic concepts, and thus can be exploited to guide the style color transfer. An interesting observation, as shown in Fig.~\ref{fig:early} (a), is that the color and illumination of the generated results are decided during the early stage of denoising of the diffusion model. To support this finding, we calculate the VGG style loss between the predicted result and the final image for different time steps under 100 examples during different time steps. As shown in Fig.~\ref{fig:early} (b), only a small difference between the initial and final stages of the denoising process (as stated in ~\cite{ke2023neural}, loss below 5 usually indicates high similarity). Therefore, we might be able to use the early, or even the first-step features to guide the color transfer.

Our method requires only one forward pass of the diffusion model in the whole process. We design an efficient priors feature mapper, mapping the stable diffusion model priors to the reference style features of our IRStyle. The mapper consists of four different convolution blocks~\cite{GAN}.
Given a random noise latent $\tilde{\mathbf{z}}_T \sim \mathcal{N}(\mathbf{0},\mathbf{I})$ at timestep $T$ and a style text prompt $\mathbf{T_s}$, we extract features from four different layers of the U-Net decoder, denoted as $\mathbf{F_t}={\{\mathbf{F_{t,i}}\}_{1\leq i\leq4}}$. 
Then, we use the mapper to transfer $\mathbf{F_t}$ to the corresponding style reference space in each scale, denoted as $\mathbf{F_m}={\{\mathbf{F_{m,i}}\}_{1\leq i\leq4}}$. We subsequently replace $\mathbf{F_s}$ in IRStyle with $\mathbf{F_m}$, for the text-guided color transfer.

\vspace{2pt}
\textbf{Training with Synthetic Data.}\quad The whole training process is shown in Fig.~\ref{fig:pipeline} (c). Since there are no public datasets that can be used for training, we design a synthetic data generation way, which is highly cost-efficient.
Firstly, we use ChatGPT to generate a style text prompt $\mathbf{T_s}$, and then feed $\mathbf{T_s}$ into the stable diffusion model to generate the corresponding style image $\mathbf{I_s}$. 
Secondly, given the content image $\mathbf{I_c}$ and $\mathbf{I_s}$, we feed them into our trained IRStyle to generate the result $\mathbf{I_g}$.
Finally, we use the $(\mathbf{I_c}, \mathbf{T_s}, \mathbf{I_g})$ as one training sample, where $\mathbf{I_c}$ and $\mathbf{T_s}$ are the inputs, and $\mathbf{I_g}$ is the ground truth. We denote the result of TRStyle as $\mathbf{Y}$. MSE loss between $\mathbf{Y}$ and $\mathbf{I_g}$ is adopted, denoted as $\mathcal{L}_{teach}$ .
During training, only the mapper is trainable and others are frozen. 

\vspace{2pt}
\textbf{Discussion.}\quad
Directly using the CLIP features to guide the color style transfer as ~\cite{radford2021learning} is another choice. The reasons behind our choice of stable diffusion features over CLIP features are discussed in the appendix.

\vspace{-4pt}
\section{Experiments}
\vspace{-1pt}
We evaluate our image style transfer in image reference setting (Sec~\ref{sec:img_compar}) and text reference setting (Sec~\ref{sec:text}). 
Evaluations on video style transfer are conducted in the appendix. 
\vspace{-1pt}
\begin{figure*}[t]
\centering
\includegraphics[width=\linewidth]{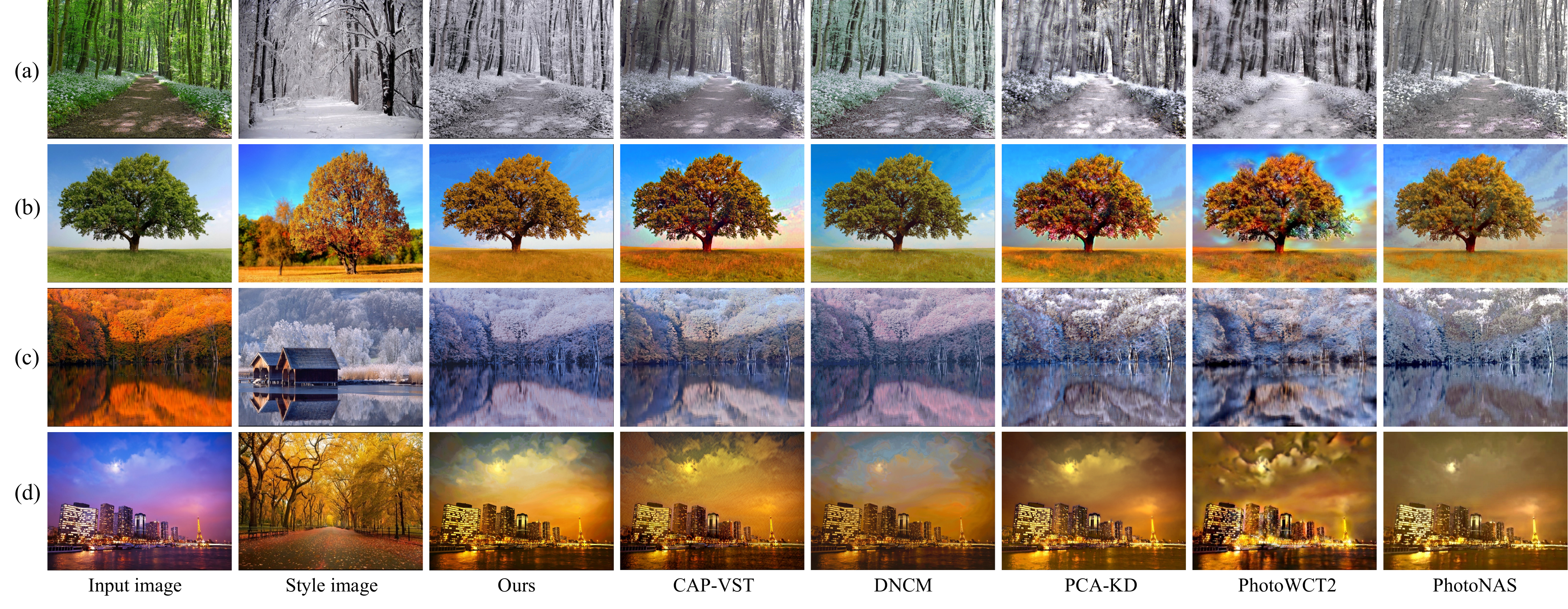}
{\begin{center}
\vspace{-8pt}
\caption{\textbf{Qualitative comparison of the image reference setting.} Our method shows the superiority in both photorealism and stylization over the other methods.
}
\label{fig:visualcomp}
\end{center}
}
\vspace{-0.7cm}
\end{figure*}

\begin{table}[t]
    \centering
    \caption{\textbf{Quantitative comparison of the image reference setting.}  The best and second best are in bold and underlined.}
    \label{tab:performance}
    \vspace{-7pt}
    \resizebox{\linewidth}{!}{
    \begin{tabular}{l|cccccc}
    \hline
        Method &Photo-NAS &PhotoWCT2 &PCA-KD  & DNCM  & CAP-VST & Ours \\ \hline
        Style Gram  loss $\downarrow$ & 2.9132 & 1.8485 & 2.2991  & 3.4972 & \textbf{0.9310} & \underline{1.2707} \\ 
        Content SSIM $\uparrow$ & 0.7132 & 0.7136 & 0.7138  & \underline{0.7777} & 0.7618 & \textbf{0.7913} \\ 
        Style score $\uparrow$& 0.8167& 0.7873 & 0.8134  & 0.7325 & \textbf{0.8647} & \underline{0.8262} \\
        User score $\uparrow$ & 2.43& 2.59 & 2.70  & 2.97 & \underline{3.35} & \textbf{3.61} \\ \hline
    \end{tabular}}
    \vspace{-7pt}
\end{table}
\vspace{-2pt}
\subsection{Image Reference Experiments}\label{sec:img_compar}
\vspace{-1pt}

\vspace{-2pt}
\subsubsection{Comparisons with Other Methods}
We compare IRStyle on the reference image color transfer task with recent methods: %
PhotoWCT2~\cite{PhotoWCT2}, PhotoNAS~\cite{an2019ultrafast}, PCA-KD~\cite{chiu2022pca}, CAP-VST~\cite{wen2023cap} DNCM~\cite{ke2023neural}. 
We use their publicly available pre-trained models and code for evaluation.
We do not compare the time cost with DNCM since it only provides online demos.

\textbf{Qualitative Results.}\quad
Fig.~\ref{fig:visualcomp} shows the visual comparison with other methods.
We can see, in most cases, PhotoWCT2, PCA-KD, and PhotoNAS exhibit significant visual noise and loss of detail. While CAP-VST mitigates this problem, it still presents visual noise (\eg, the sky in Fig.~\ref{fig:visualcomp} (d)). DNCM eliminates visual artifacts but struggles to preserve color similarity (\eg, the tree in Fig.~\ref{fig:visualcomp} (b)).
Compared with the existing methods, our method faithfully maintains image details and delivers superior stylization results. 
Besides, our method ensures consistent image stylization without artifacts, even with significant variations in color style within the input (\eg, Fig.~\ref{fig:visualcomp} (a)).

\begin{table}[t]
\begin{center}
\caption{\textbf{Comparison on GPU Inference Time\,/\,Memory, and Model Size.} 
    All evaluations are conducted with float32 model precision on a Tesla V100 GPU (32GB memory). 
    The units ``s'', ``GB'', and ``M'' refer to seconds, gigabytes, and millions, respectively.
    ``OOM'' indicates out of memory.
    }
    \vspace{-9pt}
\label{tab:speed_size}
\setlength{\tabcolsep}{4pt}
\resizebox{\linewidth}{!}{
\begin{tabular}{l|ccc}
\toprule
\textbf{Resolution} & \textbf{PhotoNAS} & \textbf{PhotoWCT2} & \textbf{PCA-KD} \\
\midrule
\textbf{FHD (1920$\times$1080)} & 0.59s / 15.6GB & 0.3s / 14GB & \underline{0.05s} / 7GB \\ 
\textbf{2K (2560$\times$1440)} & 0.99s / 23.9GB & 0.45s / 20GB & \underline{0.06s} / 11GB \\ 
\textbf{4K (3840$\times$2160)} & OOM & 1s / 23.8GB & \underline{0.1s} / 16GB \\ 
\textbf{8K (7680$\times$4320)} & OOM & OOM & OOM \\ 
\midrule
\textbf{Model Size (M)} & 40.2M & 7M & 73K \\ 
\midrule
\midrule
\textbf{Resolution} & \textbf{CAP-VST} & \textbf{DNCM} & \textbf{Ours} \\
\midrule
\textbf{FHD (1920$\times$1080)} & 1.09s / 24GB & - / \textbf{1.96GB} & \textbf{0.019s} / \underline{3GB} \\ 
\textbf{2K (2560$\times$1440)} & 1.1s / 24GB & - / \textbf{1.96GB} & \textbf{0.019s} / \underline{3GB} \\ 
\textbf{4K (3840$\times$2160)} & 1.1s / 24GB & - / \textbf{1.96GB} & \textbf{0.021s} / \underline{3GB} \\ 
\textbf{8K (7680$\times$4320)} & \underline{1.1s} / 24GB & - / \textbf{1.96GB} & \textbf{0.022s} / \underline{3GB} \\ 
\midrule
\textbf{Model Size (M)} & 4M & 5.15M & 24M \\ 
\bottomrule
\end{tabular}}
\end{center}
\vspace{-14pt}
\end{table}

\textbf{Quantitative Results.}\quad 
Following previous work~\cite{chiu2022pca, wen2023cap, ke2023neural}, we employ three metrics for evaluation, \ie, Style Gram loss~\cite{Gatys2016ImageST} and Style score~\cite{ke2023neural} to measure style similarity, and Content SSIM~\cite{ke2023neural} to measure content similarity. Table \ref{tab:performance} shows that our IRStyle provides the best trade-off between content and style similarity. Although CAP-VST achieves higher scores in style similarity, the visual artifacts however lead to worse visual effects (\eg, Fig.~\ref{fig:visualcomp} (d)). More visualizations are shown in the appendix.

\textbf{User Study.}\quad
We further conduct a user study to evaluate the subjective quality of different methods.
We invite 40 users and show them 20 randomly selected images from the test set, each consisting of an input image, a reference style image, and 6 randomly shuffled transfer results. 
Participants are requested to rate the overall stylization quality of the transfer results on a scale of 1 to 5, mainly focusing on aspects such as style and content similarity, photorealism, and the visual appeal of the color style.
After collecting these results, we calculate the average score for each method. Table \ref{tab:performance} suggests that our methods are predominantly favored by users. Detailed analyses are provided in the appendix.

\begin{table}[t]
    \centering
    \caption{\textbf{Ablation study of IRStyle.} }
    \vspace{-8pt}
    \label{tab:ablation}
    \resizebox{\linewidth}{!}{
    \begin{tabular}{l|c|c|c}
        \hline
         Type & Style Gram loss $\downarrow$& Style score $\uparrow$& Content SSIM $\uparrow$\\ \hline
         w/o interaction & 3.598 &0.7745& 0.7547 \\ 
        w/o $\mathcal{L}_{pair}$ & 2.096 & 0.8158&0.6678 \\ 
        w/o $\mathcal{L}_{unpair}$ & 1.853 & 0.8193&0.770 \\ 
        w/o $\mathcal{L}_{hist}$ & 1.385 & 0.8196&0.7720 \\ 
        w/o dual-mapping & 1.452 & \textbf{0.8294} &0.7890 \\ 
        Full version & \textbf{1.270} &0.8262& \textbf{0.7913} \\ 
        \hline
    \end{tabular}}
    \vspace{-1pt}
\end{table} 
\begin{figure}[t]
\centering
\includegraphics[width=\linewidth]{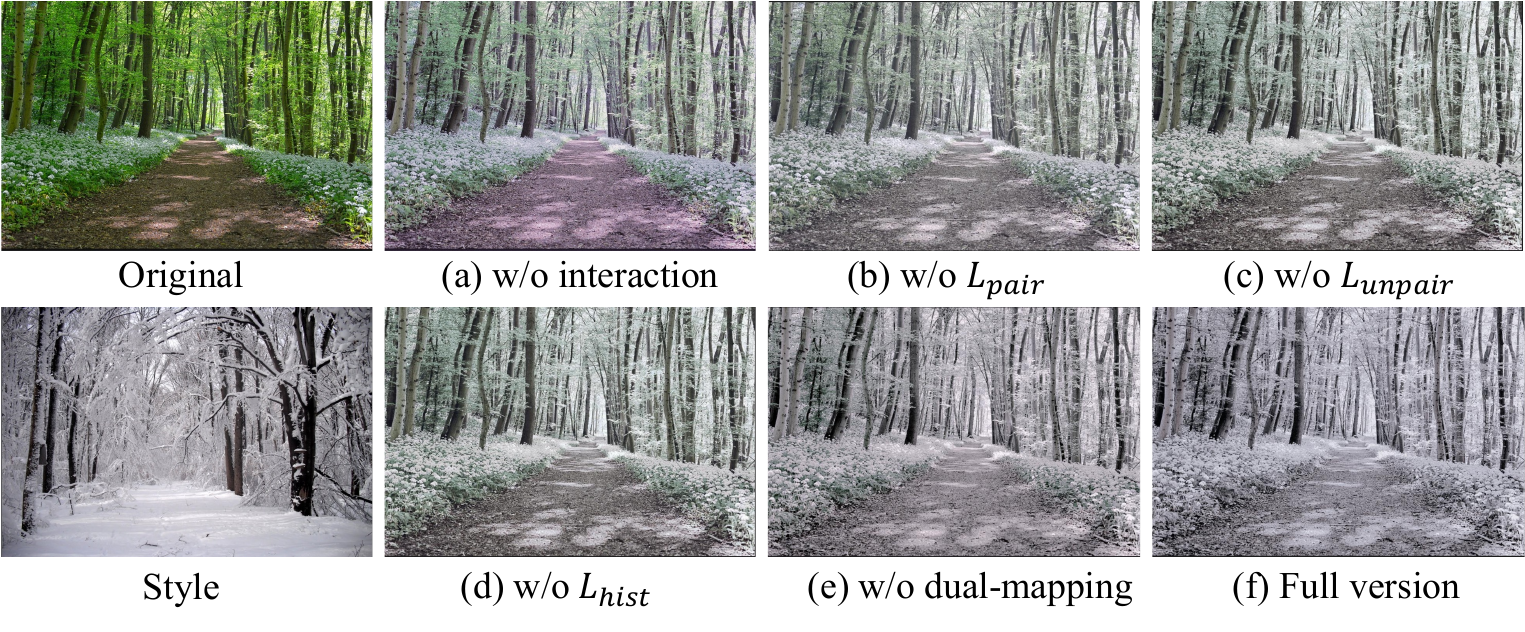}
{\begin{center}
\vspace{-0.5cm}
\caption{\textbf{Visualization results of IRStyle's ablation study.} These visual results are consistent with those in Table~\ref{tab:ablation}.
}
\label{fig:ablation}
\end{center}
}
\vspace{-15pt}
\end{figure}
\begin{figure*}[t]
\centering
\includegraphics[width=\linewidth]{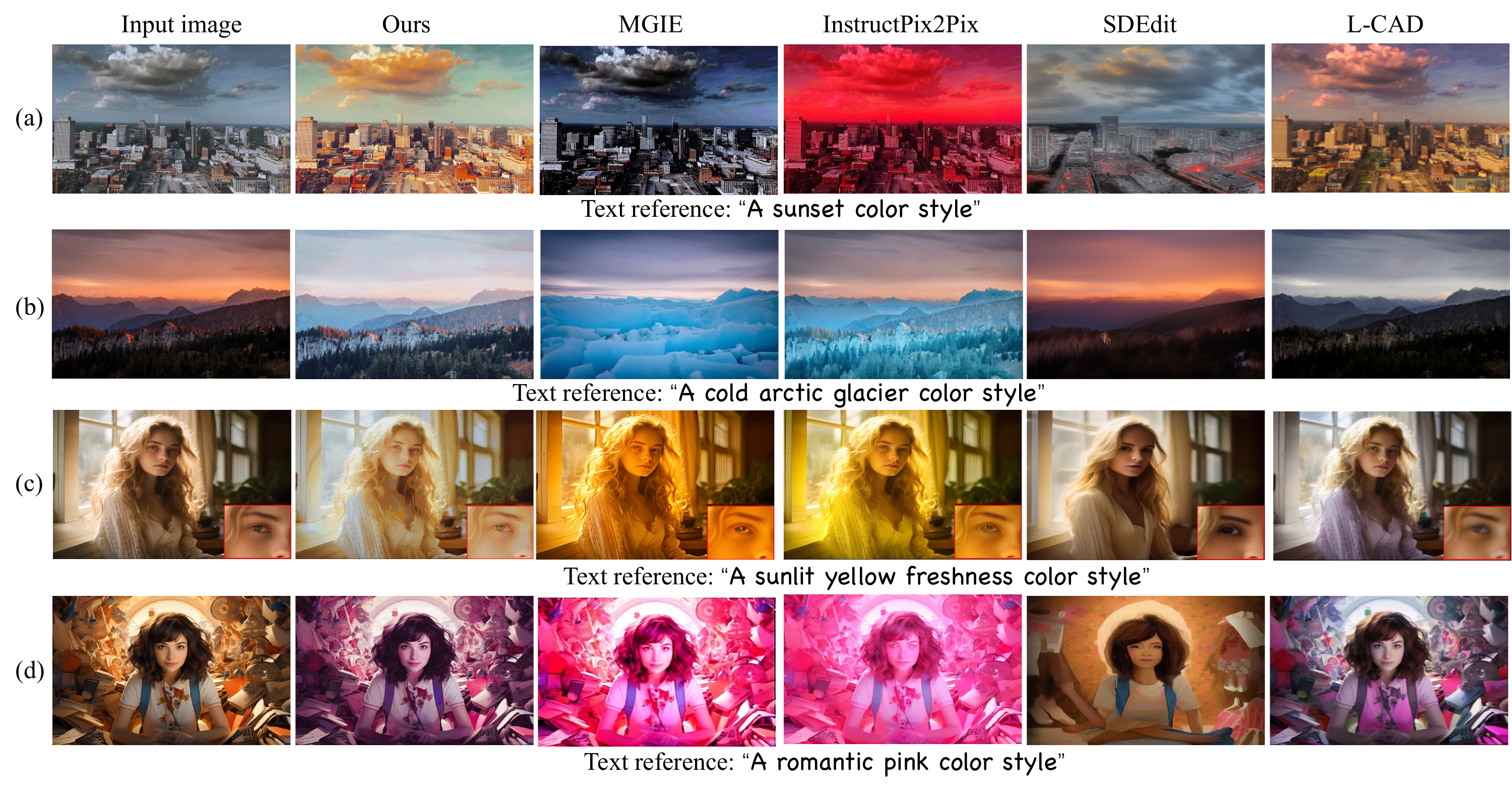}
{\begin{center}
\vspace{-16pt}
\caption{\textbf{Qualitative comparison results of text reference setting.} Our method can generate natural stylized images by open-set text reference prompts and preserve the original image texture without any finetuning.
}
\label{fig:textcomp}
\end{center}
}
\vspace{-2pt}
\end{figure*}
\textbf{Inference Time and Memory.}\quad
As shown in Table \ref{tab:speed_size}, on FHD, 2K, 4K, and 8K images, our method has the fastest inference speed in all settings, even nearly 3× speedup compared to the fastest state-of-the-art method, \ie, PCA-KD.
Moreover, the time cost of our IRStyle is essentially insensitive to practical resolutions.
Most WCT-based methods (PhotoWCT2, PhotoNAS, and PCA-KD) demand a significant amount of memory, leading to out-of-memory issues in high-resolution images, even when GPUs with 32GB of RAM are employed. In contrast, CAP-VST utilizes a uniform resize to a resolution of $1280\times 960$ to reduce the memory footprint, but at the cost of sharpness, which contradicts the purpose of using 2K and 4K images. 
DNCM performs better in terms of parameters and memory. We attribute this to the color mapping matrix, which isn't our research focus. We also implement the color transfer matrix to achieve a similar model size and competitive results.

\vspace{-6pt}
\subsubsection{Ablation Studies}\label{sec:ablation_img}
In this part, we conduct a systematic empirical study on our IRStyle.

\textbf{Interaction.}\quad 
We construct the proposed IRStyle without interaction module (Fig.~\ref{fig:dual-lut} (b)).
Table~\ref{tab:ablation} w/o interaction shows that removing the interaction module significantly decreases style similarity and slightly affects content similarity. Fig.~\ref{fig:ablation} also highlights the importance of feature interaction between the content and style images.

\textbf{Dual-mapping.}\quad
We construct the proposed IRStyle using the architecture of direct-mapping (Fig.~\ref{fig:dual-lut} (a)). Table~\ref{tab:ablation} w/o dual-mapping indicates that the dual-mapping design can enhance both style and content similarity. The visualization in Fig.~\ref{fig:ablation} further validates this conclusion.

\textbf{Supervision Functions.}\quad We validate the effectiveness of each of our proposed supervision, including $\mathcal{L}_{pair}$, $\mathcal{L}_{unpair}$ and $\mathcal{L}_{hist}$.
Both Table~\ref{tab:ablation} and Fig.~\ref{fig:ablation} confirm the significance of each supervision, \ie, employing all supervisions can yield the optimal style color transfer results.

\begin{figure}[t]
\centering
\includegraphics[width=1.0\linewidth]{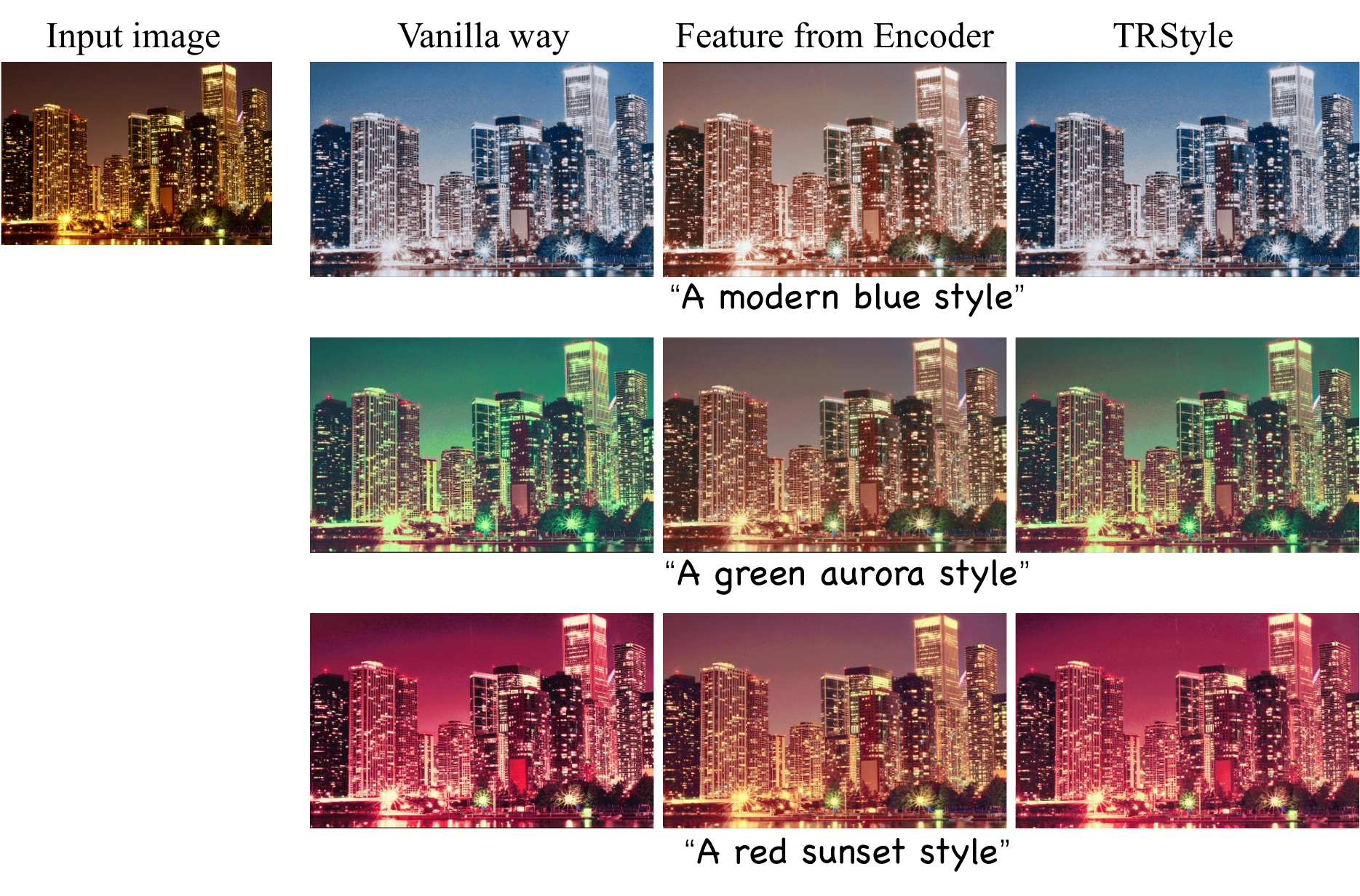}
{\begin{center}
\vspace{-1pt}
\caption{\textbf{Ablation study of TRStyle.} 
}
\label{fig:textabla}
\end{center}
}
\vspace{-3pt}
\end{figure}

\vspace{-1pt}
\subsection{Text Reference Results}\label{sec:text}
\vspace{-3pt}

\subsubsection{Comparison with Other Methods}
We compare our TRStyle with four state-of-the-art methods, including SDEdit~\cite{meng2021sdedit}, InstructPix2Pix~\cite{brooks2022instructpix2pix}, MGIE~\cite{fu2023guiding} and L-CAD~\cite{weng2024cad}. Since L-CAD falls under text-guided colorization and can only accept grayscale input, we convert the original image to grayscale for inference with L-CAD.
As shown in Fig.~\ref{fig:textcomp}, our method outperforms other techniques in terms of photorealism, stylization, and visual expressiveness. 
SDEdit exhibits content distortion and inconsistency in color style and text description. This is primarily because these methods focus on content editing and lack sufficient training for color style transfer.
While InstructPix2Pix and MGIE achieve better style consistency, it also produces images with content distortion, particularly in portrait scenes (\eg, the eyes in Fig.~\ref{fig:textcomp} (c)). Moreover, the color style is less visually attractive compared to ours. 
L-CAD exhibits no shortcomings in terms of content, however, it is relatively weaker in stylization (\eg, Fig.~\ref{fig:textcomp} (b)). This is because such methods typically concentrate on the specific colors of objects, rather than the overall color style.
More visualizations are shown in the appendix.

\textbf{Inference Time and Memory.}\quad As shown in Table \ref{tab:time}, our method surpasses others in terms of both speed and memory efficiency. We attribute this to the utilization of features derived from one single forward of the Stable Diffusion and the design of our priors feature mapper.
\begin{table}[t]
    \begin{center}
    \caption{\textbf{Comparison on Inference Time and Memory.} }
    \label{tab:time}
    \vspace{-8pt}
 \resizebox{\linewidth}{!}{
		\begin{tabular}{cccccc}
            \hline
            Method  & MGIE & InstructPix2Pix & SDEdit&L-CAD&Ours
            \\
            \hline
    Time/Memory$\downarrow$ & 13.8s/40GB&9.67/15.5GB&8.97s/15.5GB&17.6s/14GB&\textbf{0.316s}/\textbf{12.5GB} \\
			\hline
	\end{tabular}
  } 
	\end{center}
 \vspace{-2pt}
\end{table}

\vspace{-6pt}
\subsubsection{Ablation Studies} \label{sec:ablation_txt}
We provide a detailed ablation analysis of different configurations of our method.

\textbf{Compaired with the Vanilla Way.}\quad
We compare the results of the vanilla way (Sec.~\ref{sec:naive}) with our TRStyle. As shown in Fig.~\ref{fig:textabla}, our method achieves similar results to the vanilla way, but with a shorter running time (one-pass vs. $T$-pass). This demonstrates the effectiveness of our feature mapper and synthetic data collection method.

\textbf{Compaired with Feature from Encoder.}\quad We construct the proposed TRStyle with the features extracted from the encoder of the U-Net. As shown in Fig.~\ref{fig:textabla}, it is clear that ours (features extracted from the decoder) achieves greater consistency between the final result and the text prompt. The reason is that the decoder contains more information about the text prompt, as confirmed in \cite{cao2023masactrl}.

\section{Conclusion}
\vspace{-4pt}
In this paper, we propose a universal multi-modality reference color style transfer architecture named MRStyle, which accepts prompts from either images or text as references.
This is the first time that unification has been achieved in the text modality and image modality for color style transfer.
Benefiting from the proposed interaction dual-mapping network and the combined supervised learning pipeline, our method shows significant improvements over existing methods in various aspects when using image reference. 
Additionally, owing to our proposed efficient priors feature mapper and data construction methods, our method shows superiority and effectiveness when accepting text references.

\bibliography{aaai25}

\clearpage
\appendix

\twocolumn[{%
\renewcommand\twocolumn[1][]{#1}%

\vspace{-1.9cm}
\begin{center}
    \centering
    \includegraphics[width=1\linewidth]{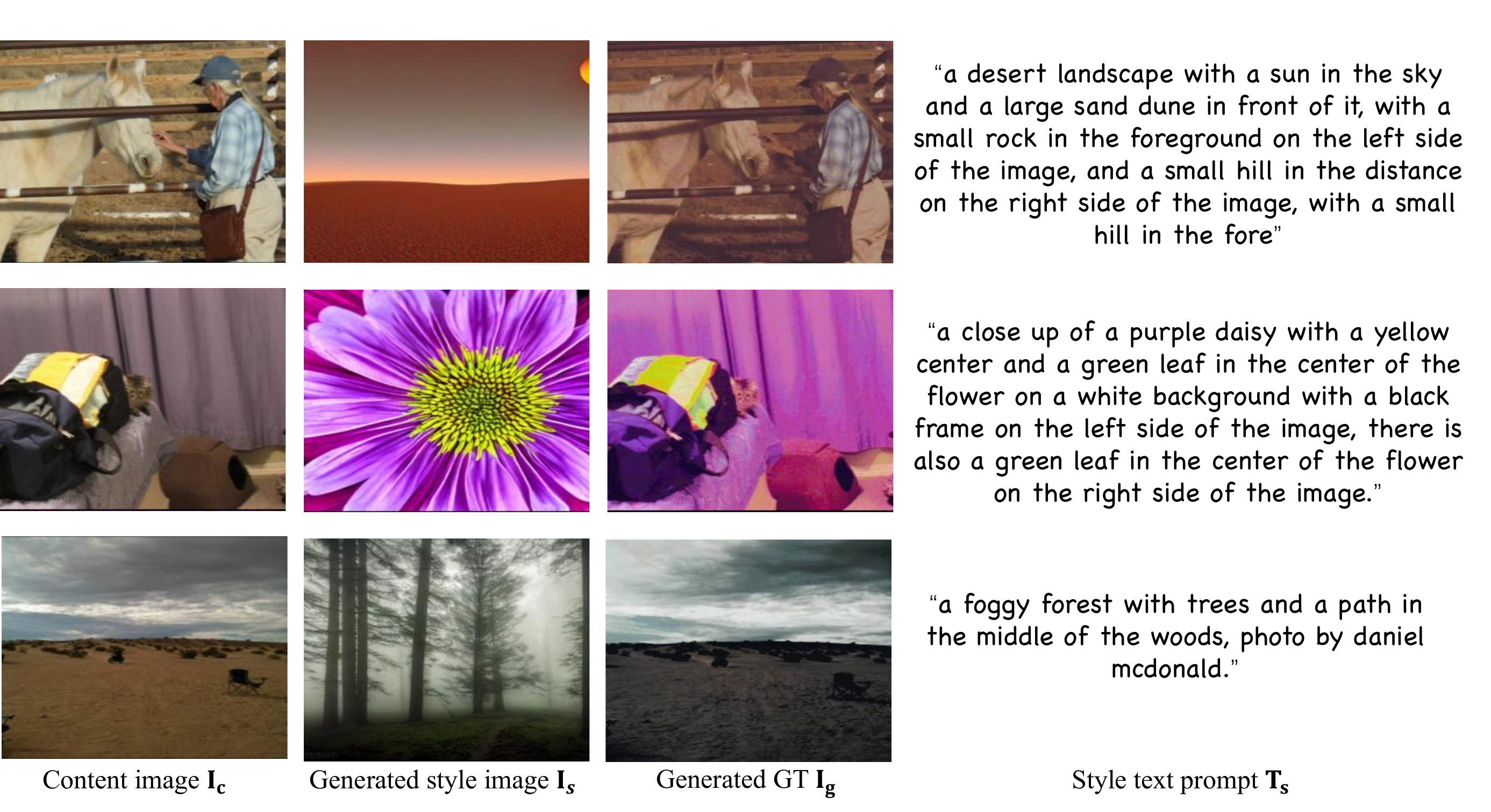}
    \vspace{-19pt}
    \captionof{figure}{
    \textbf{Samples of synthetic data for TRStyle training.
    }}
    \label{fig:datashow}
    \vspace{10pt}
\end{center}%
}]

\section{Synthetic Data for TRStyle Training}
We use ChatGPT~\cite{ouyang2022training} and Stable Diffusion Model~\cite{rombach2022high} (v1-5) to make 100000 text-image pairs. 
Each pair consists of a sentence within 70 words as the style text reference $\mathbf{T_s}$ and its corresponding generated style image $\mathbf{I_s}$. Then we randomly select content images $\mathbf{I_c}$ from the COCO~\cite{MSCOCO} dataset and construct training triplets $(\mathbf{I_c}, \mathbf{T_s}, \mathbf{I_g})$ following Sec. 3.2.2.
In Fig.~\ref{fig:datashow}, we show the samples of our synthetic data.

\section{Implementation Details}\label{sec:setting}
\subsection{Implementation of IRStyle}
Following recent color style transfer methods~\cite{WCT2,an2019ultrafast,PhotoWCT}, we train our model on the images from the MS COCO~\cite{MSCOCO} dataset.
We collect about 6,000 3D-LUT files as filters used in the paired supervised learning pipeline. 
During the evaluation, we use the test images collected by Photo-NAS~\cite{an2019ultrafast}.
We take the VGG as the encoder and utilize CLUT~\cite{zhang2022clut} as our color mapping LUT. The inputs are randomly cropped to $256 \times 256$.
We train the network by the Adam~\cite{Adam} optimizer for $300$ epochs. With a batch size of $24$, the initial learning rate is $5e^{-4}$.

\subsection{Implementation of TRStyle}\label{sec:textsetting}
We use ChatGPT~\cite{ouyang2022training} and Stable Diffusion Model~\cite{rombach2022high} (v1-5) to make a large dataset including 100,000 text-image pairs. Details about the dataset are illustrated in the appendix. We train the mapper network in TRStyle by Adam~\cite{Adam} optimizer for $200$ epochs. With a batch size of $8$, the initial learning rate is $5e^{-4}$ and is multiplied by $0.5$ after $30$ epochs.
The experiments are conducted on a single
Tesla V100 32G GPU.

\section{Performance on Video Color Style Transfer}
For image reference, we compare our method with state-of-the-art methods~\cite{wu2022ccpl, wen2023cap}. For quantitative evaluation, we gathered 25 pairs of video clips and corresponding style images from various scenes on the Internet. In line with~\cite{wu2022ccpl, wen2023cap}, we adopt the temporal loss to measure temporal consistency.

\begin{table}[t]
\centering
\caption{\textbf{Quantitative comparison of image reference video color transfer.} ’i’ denotes frame interval. The execution time is the total processing time for 150 frames.}
\resizebox{\linewidth}{!}{
\begin{tabular}{lcccccc}
\hline
\multirow{2}{*}{Method} & \multirow{2}{*}{Content SSIM$\uparrow$} & \multirow{2}{*}{Style Gram loss$\downarrow$} & \multicolumn{2}{c}{Temporal loss$\downarrow$} & \multirow{2}{*}{Time cost$\downarrow$}\\ \cline{4-5}
               &   &  & $i=1$  & $i=10$ \\ \hline
        CCPL & 0.7007 & 1.79 & 0.092 & 0.173 & \underline{19.27s}\\
        CAP-VST & \underline{0.7446} & \textbf{0.98} & \underline{0.071}& \underline{0.132} & 160s\\ 
        Ours & \textbf{0.7766} & \underline{1.48} & \textbf{0.054} &\textbf{0.108} &  \textbf{1.39s}\\ \hline
\end{tabular}}
\label{tab:video}
\end{table}

\begin{figure}[t]
\begin{center}
    \centering
    \includegraphics[width=1\linewidth]{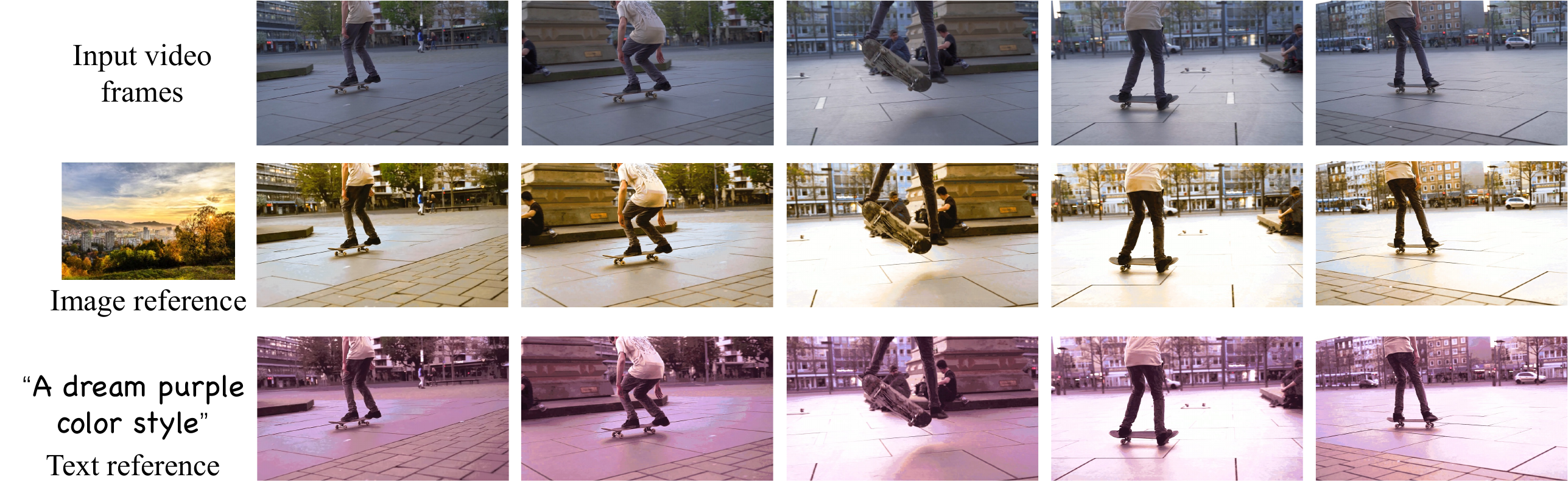}
    \vspace{-19pt}
    \captionof{figure}{
    \textbf{Video color style transfer results of MRStyle} }
    \label{fig:video_result}
\end{center}%
\end{figure}

For each video, we generate the LUT using IRStyle on the initial frame and apply it to all frames, ensuring rapid execution speed and consistent results across frames.
The results in Table~\ref{tab:video} show that our framework yields comparable results against the other methods, including temporal consistency, content similarity, style effect, and time cost. In practical scenarios, we can employ a simple scene judgment method, such as the lab histogram~\cite{sergyan2008color}, to segment video scenes. Subsequently, using the same LUT within each scene enables us to achieve the best trade-off between temporal consistency and style effect.

In Fig.~\ref{fig:video_result}, we present a series of stylized frames from our MRStyle, including image and text reference. The style remains consistent across frames and the resulting video is notably stable.

\section{User Study of IRStyle}

\begin{figure}[t]
\centering
\includegraphics[width=0.99\linewidth]{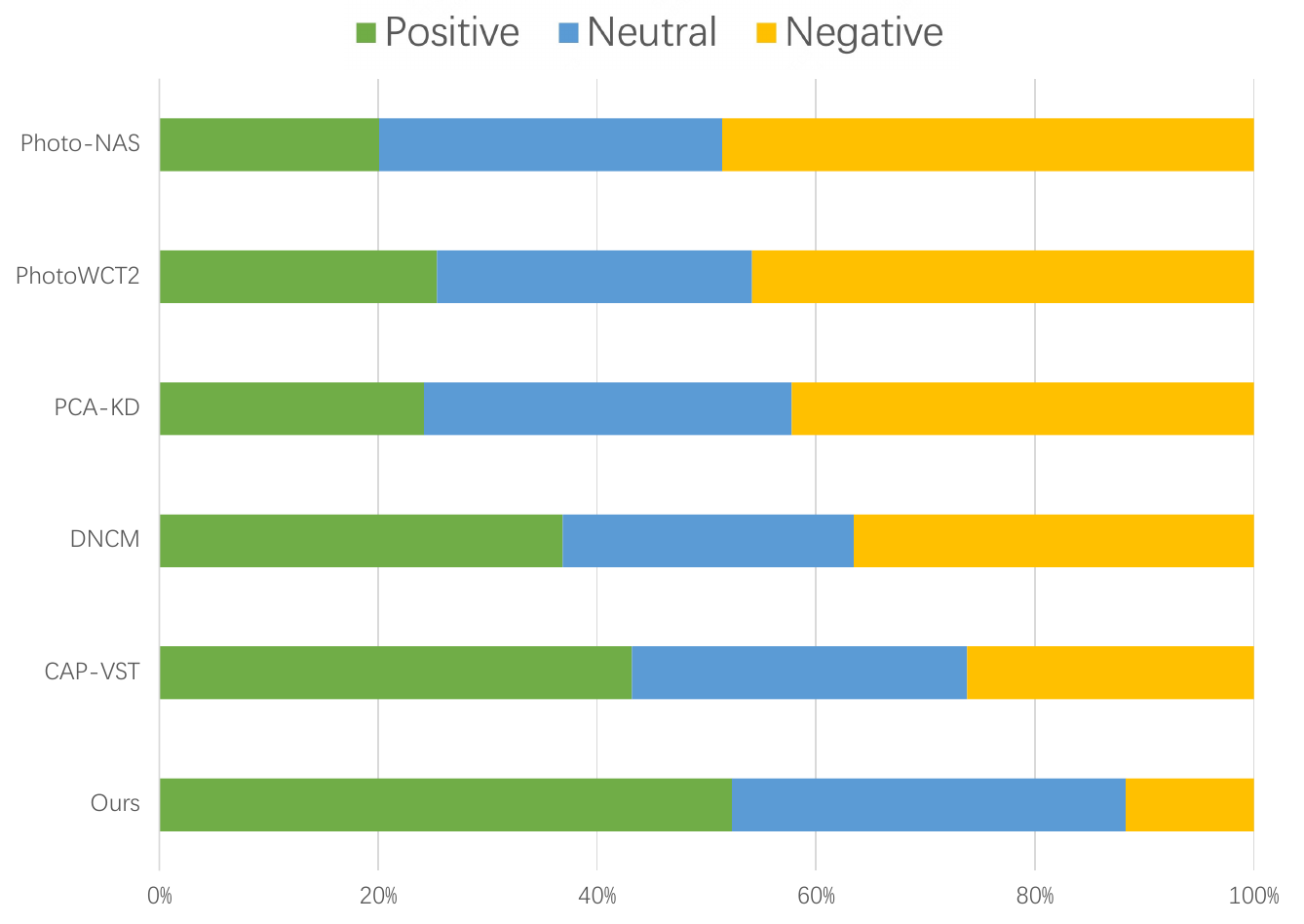}
{\begin{center}
\vspace{-0.3cm}
\caption{\textbf{Comparison on user study results.}
}
\label{fig:user-tab}
\end{center}
}
\vspace{-0.2cm}
\end{figure}

\begin{table}[!h]
    \centering
    \caption{\textbf{Quantitative comparison of the text reference setting.}  The best and second best are in bold and underlined, respectively.}
    \label{tab:text_pefor}
    \resizebox{\linewidth}{!}{
    \begin{tabular}{l|ccccc}
    \hline
        Method &MGIE\quad\quad &InstructPix2Pix\quad\quad &SDEdit\quad\quad  & L-CAD\quad\quad & Ours \\ \hline
        CLIP Score $\uparrow$  & 0.2448 & \textbf{0.2731} & 0.2366 & 0.2264&\underline{0.2466}\\ 
        Content SSIM $\uparrow$ & 0.6796 & 0.6698  & 0.4111 & \underline{0.7508} & \textbf{0.7527} \\ 
        LPIPS $\downarrow$ & 0.6796 & 0.3212  & 0.3612 & \underline{0.2531} & \textbf{0.2138} \\ 
        User score $\uparrow$ & 2.89 & \underline{3.46}  & 2.57 & 3.05 & \textbf{3.72} \\ \hline
    \end{tabular}}
\end{table}

\begin{figure}[t]
\centering
\includegraphics[width=0.99\linewidth]{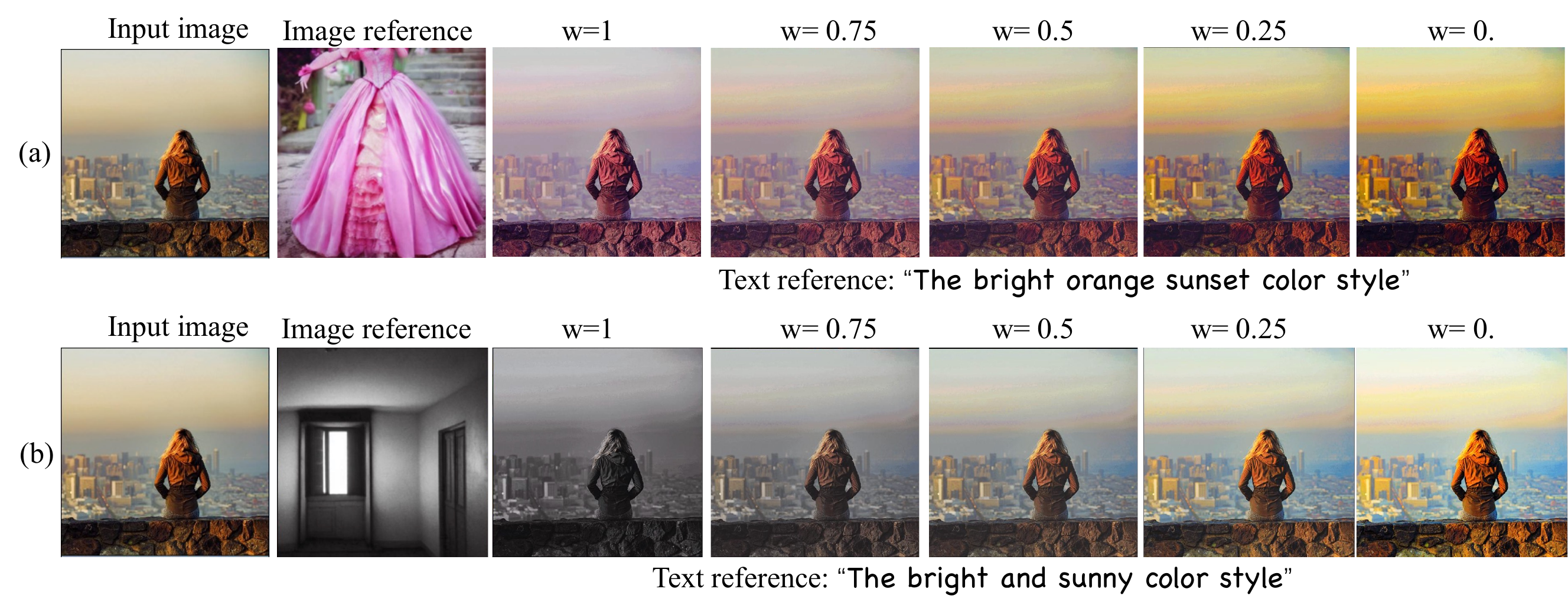}
{\begin{center}
\vspace{-0.3cm}
\caption{\textbf{Extending MRStyle to support simultaneous input of text and images.} $\mathbf{w}$ is the weight of image reference features. When $\mathbf{w}$ equals 1, only the image reference is considered, while a value of 0 represents exclusive reliance on the text reference. (a) The row emphasizes the gradual change effect between various color series (\eg, pink and orange). (b) The row depicts variations in brightness levels.}
\label{fig:miximg}
\end{center}
}
\vspace{-0.5cm}
\end{figure}
\begin{figure*}[h!]
\centering
\includegraphics[width=\linewidth]{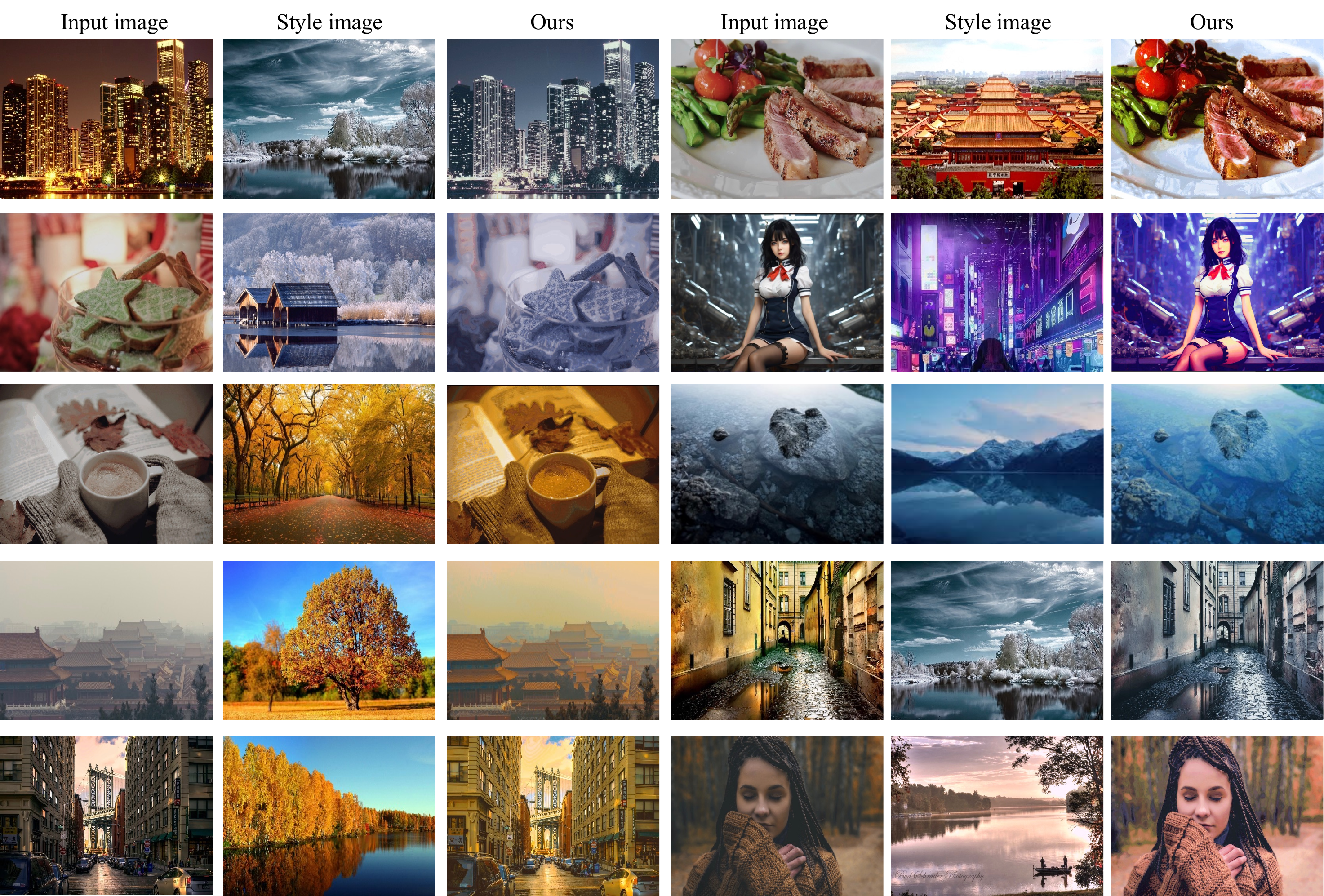}
{\begin{center}
\vspace{-0.3cm}
\caption{\textbf{Visual results of IRStyle.} Our method is robust when generalizing to different input colors.
}
\label{fig:visualimg}
\end{center}
}
\vspace{-15pt}
\end{figure*}
The overall stylization quality score is rated from 1 (least satisfactory) to 5 (highly satisfactory), with 3 indicating acceptable results. Ratings of 1 and 2 denote negative outcomes, 3 signifies neutral outcomes, while 4 and 5 represent positive outcomes. Fig.~\ref{fig:user-tab} displays the evaluation results' distribution for each method. The transfer results of our method exhibit the highest proportion in the range from neutral to positive, thus suggesting its effectiveness across a diverse range of cases.

\section{Quantitative Experiments of TRStyle}

Since there is presently no standardized benchmark for evaluating the color style transfer of text references, we have created a test benchmark consisting of 40 samples. Each sample within this benchmark includes an input image and a style text prompt. The images for the benchmark have been sourced from Unsplash\footnote{https://unsplash.com/} and civitai\footnote{https://civitai.com/}, covering four distinct scenes: buildings, sky, portraits, and mountains. The style text prompts are generated using ChatGPT~\cite{ouyang2022training}, \eg, "A gray tone, elegant and vintage style".
Following the experiments setting of IRStyle, we employ 4 metrics for evaluation of IRStyle, \ie, CLIP Score loss~\cite{hessel2021clipscore} to measure style similarity, Content SSIM~\cite{ke2023neural} and LPIPS~\cite{zhang2018unreasonable} to measure content similarity, and user score for human sensory evaluation. Table \ref{tab:text_pefor} shows that our TRStyle achieves the best trade-off between the content and style similarity, aligning most closely with popular preferences.
Our approach did not yield the highest results in terms of the clip score. This can be attributed to the fact that image editing methods often modify the content of the image, leading to a higher clip score.

\begin{figure*}[t]
\centering
\includegraphics[width=\linewidth]{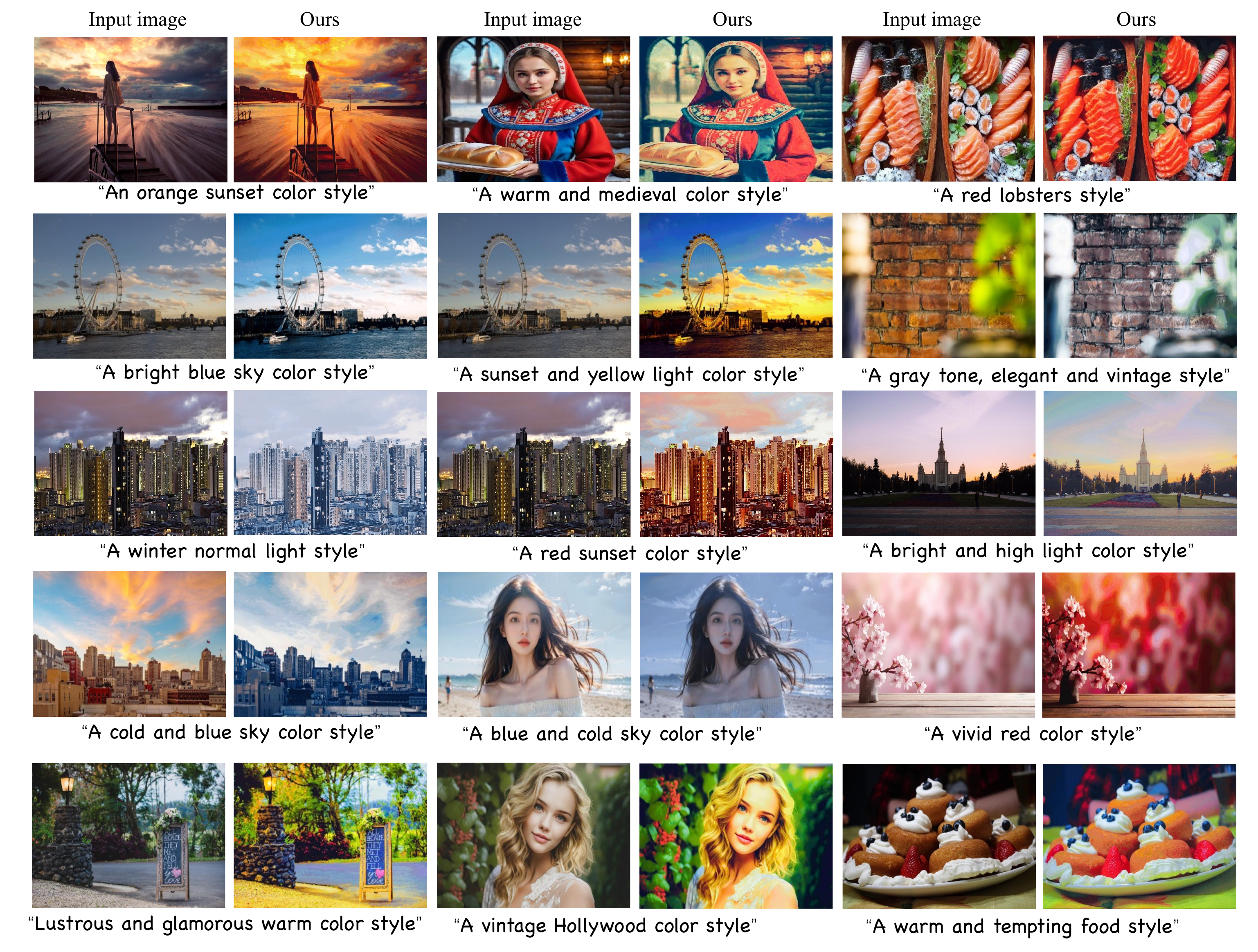}
{\begin{center}
\vspace{-0.3cm}
\caption{\textbf{Visual results of TRStyle.} Our method is robust when generalizing to different scenes, \eg, portraits, landscapes, and food.
}
\label{fig:visualtext}
\end{center}
}
\vspace{-0.5cm}
\end{figure*}

\begin{figure}[t]
\centering
\includegraphics[width=1.0\linewidth]{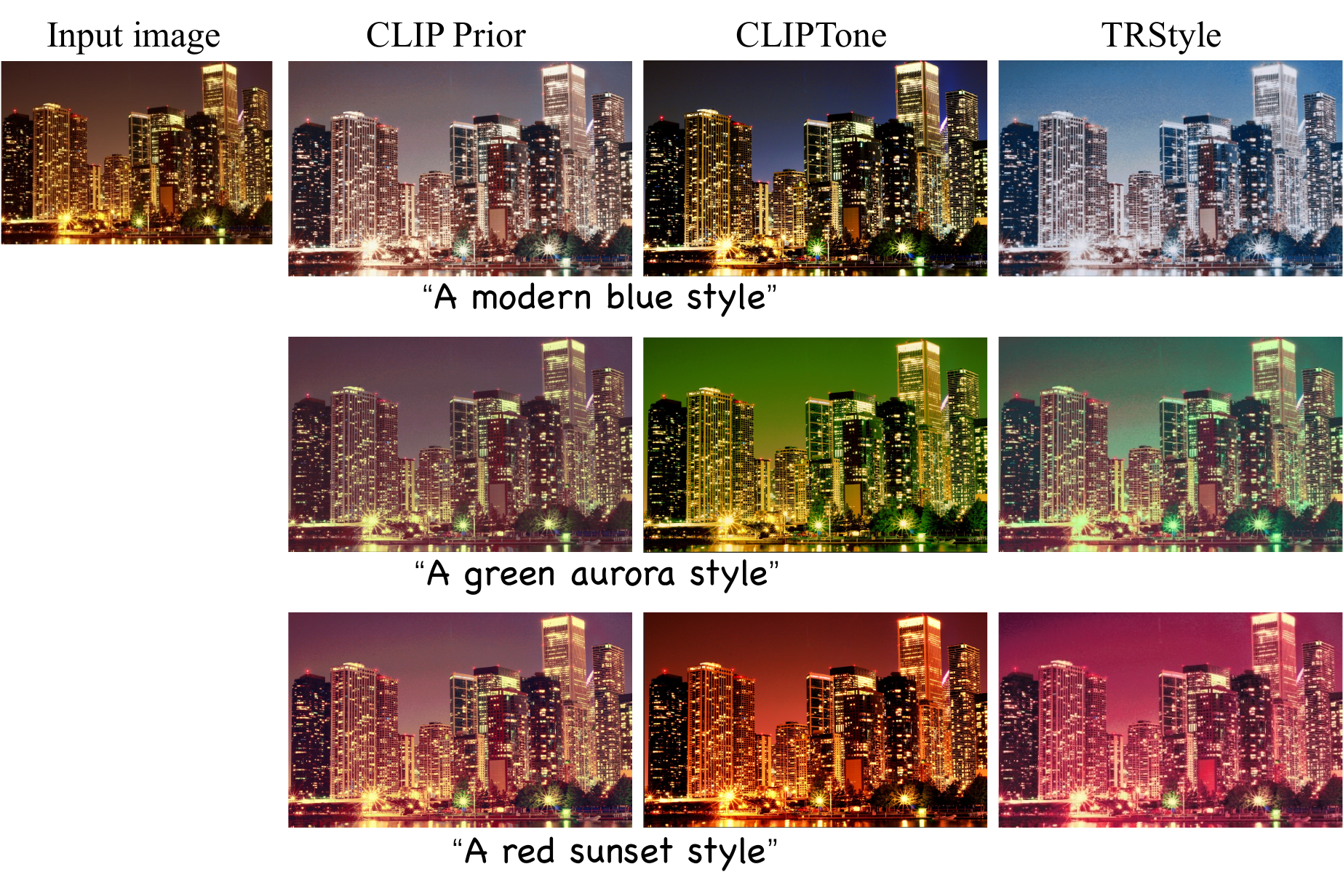}
{\begin{center}
\vspace{-17pt}
\caption{\textbf{Dissuasion with CLIP Prior in TRStyle.} 
}
\label{fig:textclip}
\end{center}
}
\vspace{-13pt}
\end{figure}

\section{Dissuasion with CLIP Prior}

We chose stable diffusion features over CLIP features for three primary reasons. First, our IRStyle operates in image space, which aligns more naturally with the SD’s features due to their lower-level, pixel-wise representation. This contrasts with CLIP's inclination towards capturing abstract, high-level information, which lacks the granularity of pixel-wise guidance. 
Second, the powerful image generation capability of SD can provide aesthetic priors for the final result, leading to better visual effects. 
Third, the same style description often has different tones, which is consistent with the randomness of the diffusion model. In contrast, CLIP tends to establish a fixed-style correspondence per prompt, which restricts variability. 

We further replace the SD prior in our proposed TRStyle with the CLIP prior, which involves directly extracting text features from CLIP.
As depicted in Fig.~\ref{fig:textclip}, the CLIP Prior setting produces bad transfer results. This further confirms the first reason that, compared to CLIP, the features from the SD align better with the style features of IRStyle, as both operate at the image level. We posit that a carefully developed mapper like CLIPTone~\cite{lee2024cliptone}, integrating the directional color vector, could potentially alleviate this issue. As shown in Fig.~\ref{fig:textclip}, the CLIPTone shows better text consistency compared to our CLIP Prior method. 
However, it seems that CLIPTone mainly concentrates on the color terms (\eg, 'blue' and 'green'), but overlooks the context related to these colors (\eg, 'modern' and 'aurora').
Moreover, when compared to our SD prior method (TRStyle), its color tones appear overly saturated. This aligns with our earlier analysis that the SD-based approach can leverage the aesthetic priors of SD, thus yielding better visual results.

\section{Simultaneous Text and Image References}
We extend our MRStyle to support simultaneous text and image references. The style feature of text and image references are denoted as $\mathbf{F_t}$ and $\mathbf{F_i}$ respectively.
Since we align the style information in the text and image into a common space, we can mix these styles by $\mathbf{F_m} = \mathbf{w}*\mathbf{F_i} + (1-\mathbf{w})*\mathbf{F_t}$, where $\mathbf{w} \in [0, 1]$.
Subsequently, we utilize the $\mathbf{F_m}$ as the style feature in our interaction dual-mapping network to complete the style transfer. 
Fig. \ref{fig:miximg} presents the depiction of our method’s capability to simultaneously support reference inputs of text and images.
By adjusting the weight $\mathbf{w}$, we can effectively balance the contribution of both modes.
As $\mathbf{w}$ increases, the transfer becomes more biased towards the style indicated by the image references.
In the event of substantial disparities between textual and pictorial references, our methodology may result in unconventional or unexpected outcomes. However, due to the scope and limitations of this paper, a comprehensive analysis of this phenomenon will be excluded from the detailed discussion provided.

\begin{figure}[t]
\centering
\includegraphics[width=0.99\linewidth]{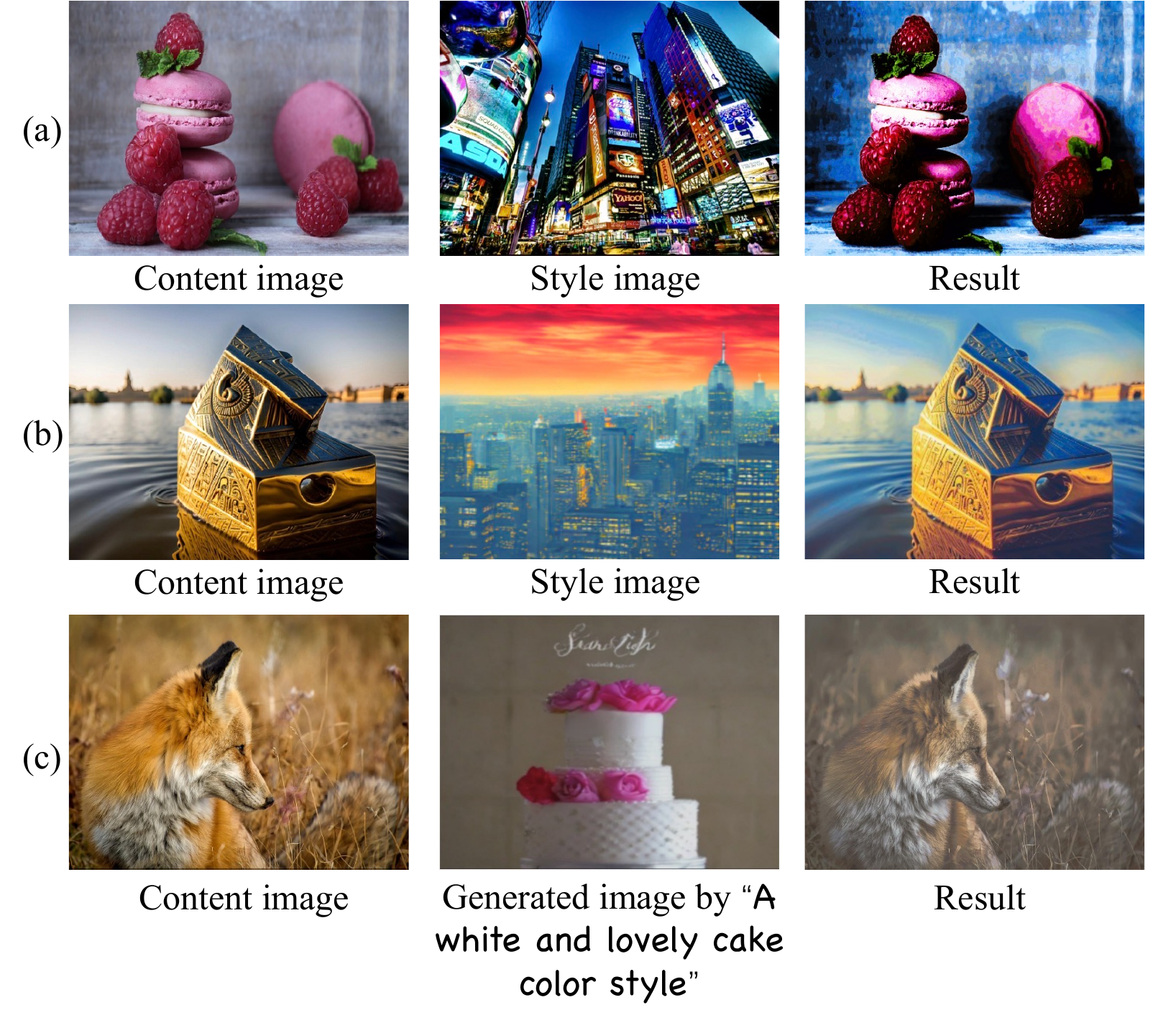}
{\begin{center}
\vspace{-0.3cm}
\caption{\textbf{Visual results of failure cases.}
}
\label{fig:limitation}
\end{center}
}
\vspace{-0.5cm}
\end{figure}
\section{Limitation} \label{sec:limitation}
For IRStyle, blurriness present in the input could potentially be amplified in the output, as shown in Fig.~\ref{fig:limitation} (a), where the JPEG artifacts are amplified, especially in the background. 
Moreover, it cannot perform local color adjustments, with Fig.~\ref{fig:limitation} (b) illustrating its inability to distinguish elements with similar colors but different semantics, such as the sky and the river.
For text reference, since our TRStyle utilizes the priors of IRStyle and the stable diffusion, the limitation mentioned above of IRStyle exists in the results of TRStyle as well. Furthermore, the style similarity will be influenced by the generation ability of the Stable Diffusion. As shown in  Fig.~\ref{fig:limitation} (c), its transfer result is inconsistent with the text description, this problem is caused by the stable diffusion, which produces a brown-style reference image.

\section{More Results}

\subsection{Visual Results of IRStyle}
We provide more visual results of IRStyle in Fig.~\ref{fig:visualimg}. 

\subsection{Visual Results of TRStyle}
\begin{figure}[t]
\centering
\includegraphics[width=0.99\linewidth]{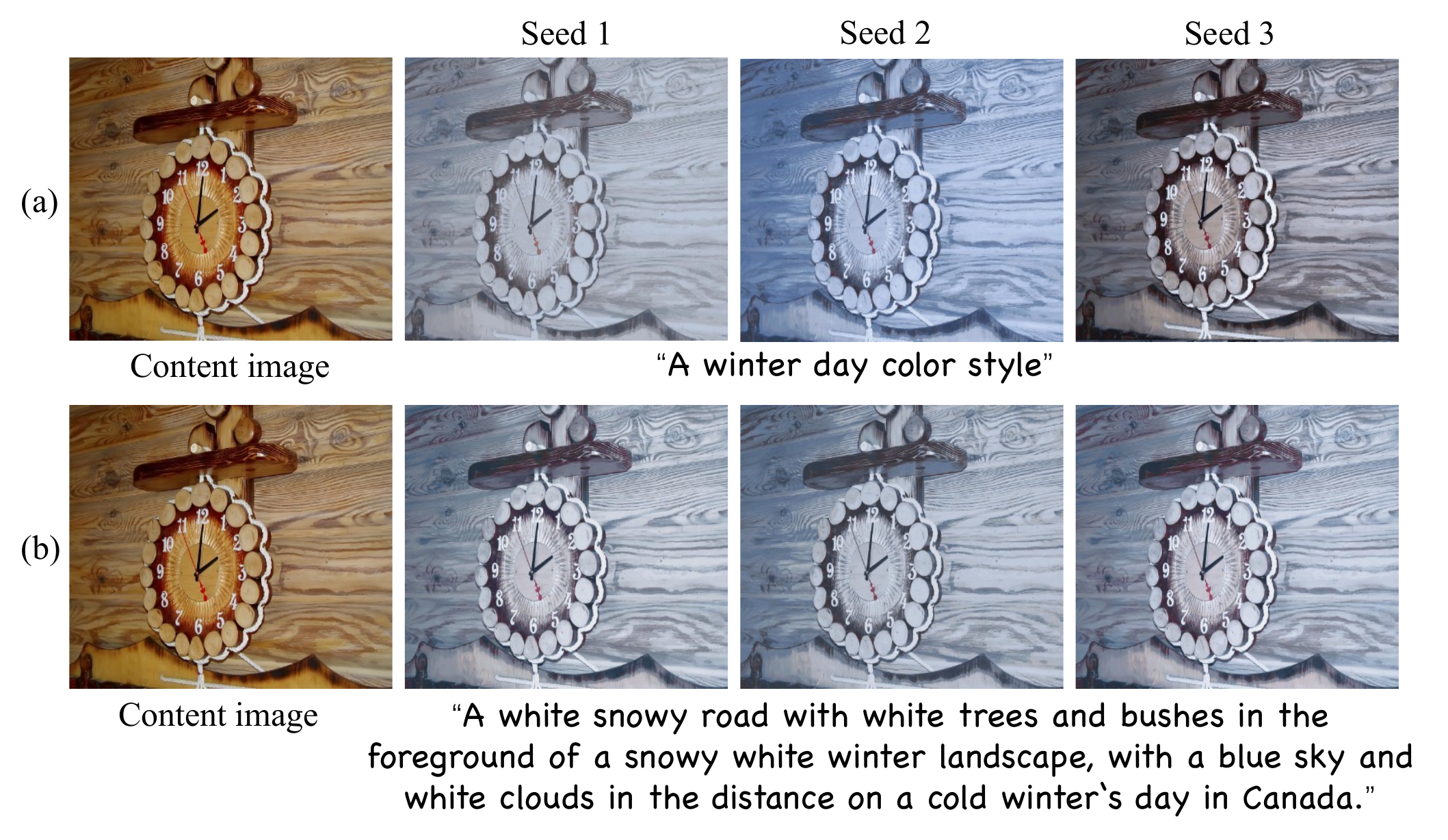}
{\begin{center}
\vspace{-0.3cm}
\caption{\textbf{Visual results of TRStyle with different noise seeds.} (a) Due to the stochasticity of the stable diffusion model, our TRStyle can produce varying color style transformation outputs for a single simple text prompt, using different noise seeds.
(b) Providing detailed text prompts, our TRStyle can generate similar color outputs for different noise seeds.
}
\label{fig:stochasticity}
\end{center}
}
\vspace{-0.5cm}
\end{figure}

We provide more visual results of TRStyle in Fig.~\ref{fig:visualtext}. 
In reality, a single scene can correspond to various color styles. For instance, a winter scene can be associated with a snowy white color style or vintage tones. Existing image editing software usually provides various styles of filters for the same scene (\eg, Winter 1, Winter 2). Leveraging the stable diffusion priors, our method can generate multiple styles for the same scene using different noise seeds with simple descriptions, as depicted in Fig.~\ref{fig:stochasticity} (a). Our approach also enables detailed descriptions of color styles, which can reduce generation ambiguity as shown in Fig.~\ref{fig:stochasticity} (b).

\end{document}